%% file: main.tex
\documentclass[10pt,twocolumn,letterpaper]{article}

\usepackage[pagenumbers]{cvpr} %

\usepackage{graphicx, nicefrac}
\usepackage{amsmath}
\usepackage{amssymb}
\usepackage{booktabs}
\usepackage{multirow}
\usepackage[table,xcdraw]{xcolor}
\usepackage[margin=0pt,position=bottom,font=footnotesize,labelfont=bf]{caption}

\usepackage[accsupp]{axessibility}  %

\usepackage[pagebackref,breaklinks,colorlinks]{hyperref}

\usepackage{cuted}

\usepackage[capitalize]{cleveref}
\crefname{section}{Sec.}{Secs.}
\Crefname{section}{Section}{Sections}
\Crefname{table}{Table}{Tables}
\crefname{table}{Tab.}{Tabs.}

\usepackage{dblfloatfix}
\usepackage{float}
\usepackage{pifont}%
\newcommand{\cmark}{\ding{51}}%
\newcommand{\xmark}{\ding{55}}%

\DeclareMathOperator*{\argmax}{\bf arg\,max}
\DeclareMathOperator*{\argmin}{\bf arg\,min}

\newcommand{\vv}{\ensuremath{\mathbf v}}

\newcommand{\squishlist}{
   \begin{list}{$\bullet$}
    { \setlength{\itemsep}{0pt}      \setlength{\parsep}{3pt}
      \setlength{\topsep}{3pt}       \setlength{\partopsep}{0pt}
      \setlength{\leftmargin}{1.0em} \setlength{\labelwidth}{1em}
      \setlength{\labelsep}{0.5em} } }
      
\newcommand{\squishend}{
    \end{list}  }

\def\cvprPaperID{7059} %
\def\confName{CVPR}
\def\confYear{2023}

\begin{document}

\title{
Ensemble-based Blackbox Attacks on Dense Prediction 
}

\author{Zikui Cai$^*$, Yaoteng Tan$^*$, and M. Salman Asif\\
University of California Riverside\\
{\tt\small \{zcai032,ytan073,sasif\}@ucr.edu}
}
\maketitle

\def\thefootnote{*}\footnotetext{Equal contribution}

\begin{abstract}
We propose an approach for adversarial attacks on dense prediction models (such as object detectors and segmentation). It is well known that the attacks generated by a single surrogate model do not transfer to arbitrary (blackbox) victim models. Furthermore, targeted attacks are often more challenging than the untargeted attacks. In this paper, we show that a carefully designed ensemble can create effective attacks for a number of victim models. In particular, we show that normalization of the weights for individual models plays a critical role in the success of the attacks. We then demonstrate that by adjusting the weights of the ensemble according to the victim model can further improve the performance of the attacks. We performed a number of experiments for object detectors and segmentation to highlight the significance of the our proposed methods. Our proposed ensemble-based method outperforms existing blackbox attack methods for object detection and segmentation. Finally we show that our proposed method can also generate a single perturbation that can fool multiple blackbox detection and segmentation models simultaneously. Code is available at \href{https://github.com/CSIPlab/EBAD}{\texttt{https://github.com/CSIPlab/EBAD}}.

\end{abstract}

\section{Introduction}
\label{sec:intro}

Computer vision models (e.g., classification, object detection, segmentation, and depth estimation) are known to be vulnerable to carefully crafted adversarial examples~\cite{szegedy2013intriguing,goodfellow2014explaining,cai2022context,gu2021adversarial,cheng2022physical}. 
Creating such adversarial attacks is easy for whitebox models, where the victim model is completely known~\cite{goodfellow2014explaining,kurakin2016adversarial,madry2017towards,dong2018boosting,xie2019improving}. In contrast, creating adversarial attacks for blackbox models, where the victim model is unknown, remains a challenging task \cite{liu2017delving,xie2017adversarial,arnab2018robustness}. Most of the existing blackbox attack methods have been developed for classification models \cite{lord2022attacking,huang2019black,cheng2019improving,tashiro2020diversity}. Blackbox attacks for dense prediction models such as object detection and segmentation are relatively less studied \cite{cai2022context, gu2021adversarial, liang2022large}, and most of the existing ones mainly focus on untargeted attacks \cite{gu2021adversarial}. 
Furthermore, a vast majority of these methods are based on transfer attacks, in which a surrogate (whitebox) model is used to generate the adversarial example that is tested on the victim model. However, the success rate of such transfer-based attacks is often low, especially for targeted attacks \cite{huang2019black,cheng2019improving,tashiro2020diversity}.

\begin{figure}[t]
    \centering
    \includegraphics[width=0.45\textwidth]{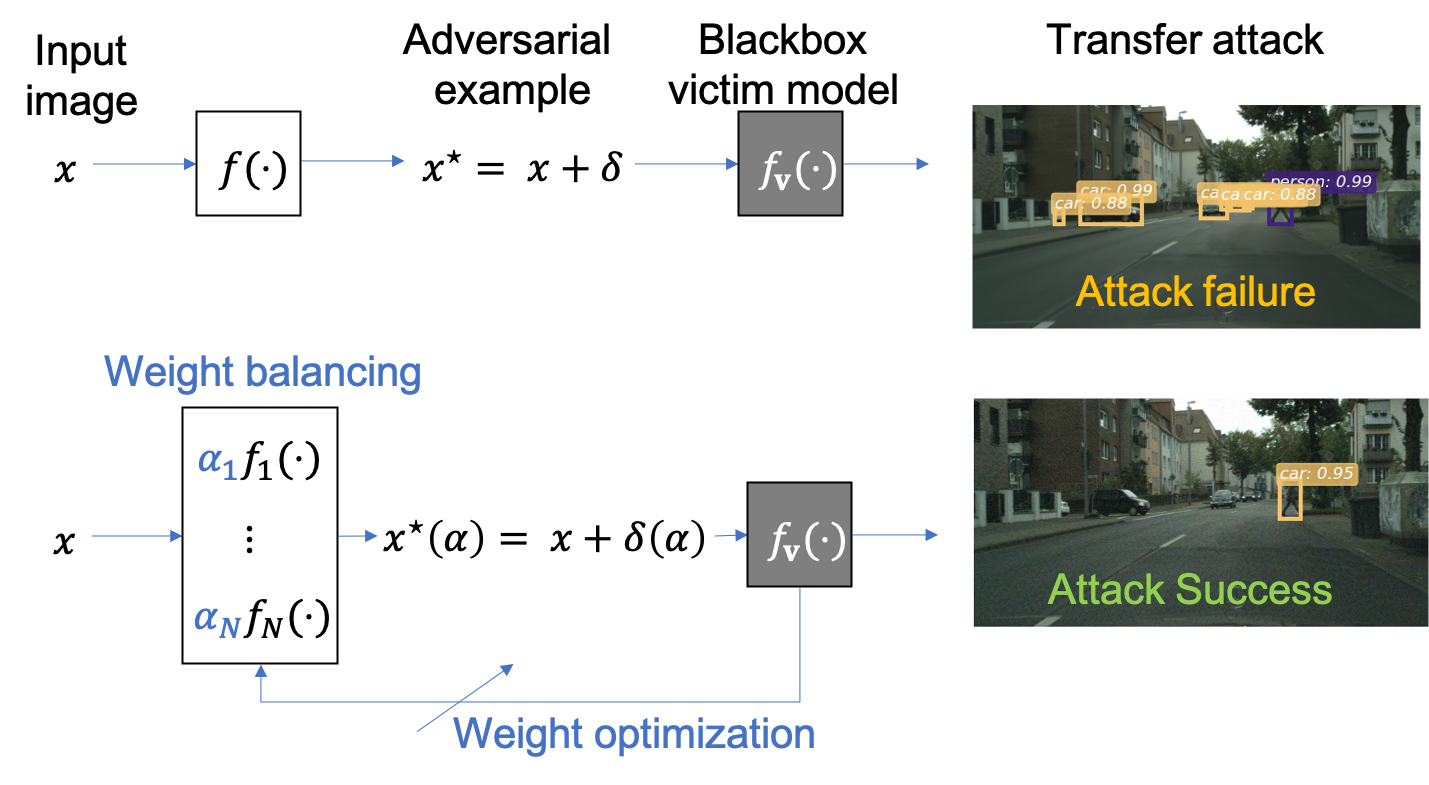}
    \caption{Illustration of the targeted ensemble-based blackbox attack. (Top) Attack generated by a single surrogate model does not transfer on the victim blackbox model (person does not map to car). (Bottom) Attack generated by weight balancing and optimization can transfer on a variety of victim models (person is mapped to car).}\label{fig:intro}
    \vspace{-5mm}
\end{figure}

In this paper, we propose and evaluate an ensemble-based blackbox attack method for objection detection and segmentation. 
Our method is inspired by three key observations: 
1) targeted attacks generated by a single surrogate model are rarely successful; 
2) attacks generated by an ensemble of surrogate models are highly successful if the contribution from all the models is properly normalized; and 
3) attacks generated by an ensemble for a specific victim model can be further improved by adjusting the contributions of different surrogate models. 
The overall idea of the proposed work is illustrated in \cref{fig:intro}. 
Our proposed method can be viewed as a combination of transfer- and query-based attacks, where we can adjust the contribution based on the feedback from the victim model using a small number of queries (5--20 in our experiments).
In contrast, conventional query-based attacks require hundreds or thousands of queries from the victim model \cite{chen2017zoo,ilyas2018black,tu2019autozoom,guo2019simple}.

We conduct comprehensive experiments to validate our proposed method and achieve state-of-the-art performance for both targeted and untargeted blackbox attacks on object detection. 
Specifically, our proposed method attains 29--53\% success rate using only 5 queries for targeted attacks on object detectors, whereas the current state-of-the-art method \cite{cai2022context} achieves 20--39\% success rate with the same number of queries. 
Furthermore, we extend our evaluation to untargeted and targeted attacks on blackbox semantic segmentation models. Our method achieves 0.9--1.55\% mIoU for untargeted and 69--95\% pixel-wise success for targeted attacks. By comparison, the current state-of-the-art method \cite{gu2021adversarial} obtains
 0.6--7.97\% mIoU for untargeted attacks and does not report results for targeted attacks. 
To the best of our knowledge, our work is the first approach for targeted and query-based attacks for semantic segmentation.

Below we summarize main contributions of this work. 

\squishlist
\item We design a novel framework that can effectively attack blackbox dense prediction models based on an ensemble of surrogate models. %
\item We propose two simple yet highly effective ideas, namely weight balancing and weight optimization, with which we can achieve significantly better attack performance compared to existing methods. 
\item We extensively evaluate our method for targeted and untargeted attacks on object detection and semantic segmentation models and achieve state-of-the-art results. 
\item We demonstrate that our proposed method can generate a single perturbation that can fool multiple blackbox detection and segmentation models simultaneously. 
\squishend

\input{related}

\input{method}

\input{experiment}

\vspace{-5pt}
\section{Conclusion}
We presented a new method to generate targeted attacks for dense prediction task (e.g., object detectors and semantic segmentation) using an ensemble of surrogate models. We demonstrate that (victim model-agnostic) weight balancing and (victim model-specific) weight optimization play a critical role in the success of attacks. We present an extensive set of experiments to demonstrate the performance of our method with different models and datasets. Finally, we show that our approach can create adversarial examples to fool multiple blackbox models and tasks jointly. 

\noindent\textbf{Limitations.}
Our method employs an ensemble of surrogate models to generate attacks, which inevitably incurs higher memory and computational overhead. Moreover, the success of our method hinges on the availability of a diverse set of surrogate models, which could potentially limit its efficacy if such models are not readily obtainable.

\noindent\textbf{Acknowledgments.} 
This material is based upon work supported by AFOSR award FA9550-21-1-0330, NSF award 2046293, UC Regents Faculty Development grant. We acknowledge the computing support from Nautilus PRP.

{\small
\bibliographystyle{ieee_fullname}
\bibliography{refs}
}

\newpage

\input{supp}

\end{document}

%% file: related.tex
\section{Related work}

\noindent \textbf{Blackbox adversarial attacks.} In the context of blackbox attacks, the attacker cannot access the model parameters or compute the gradient via backpropagation. Blackbox attack methods can be broadly divided into two groups: transfer-based \cite{papernot2016transferability,papernot2017practical,liu2017delving,li2020towards} and query-based attacks \cite{chen2017zoo,ilyas2018black,tu2019autozoom}.
Transfer-based attacks rely on the assumption that surrogate models share similarities with the victim model, such that an adversarial example generated for the surrogate model can also fool the victim model.
Query-based methods generate attacks by searching the adversarial examples space based on the feedback obtained from the victim model through queries. They can often achieve higher success rate but may require a large number of queries.

\noindent \textbf{Ensemble-based attacks. }
Ensemble-based attacks leverage the idea of transfer attack and assume that if an adversarial example can fool multiple models simultaneously, the chances of fooling an unseen model are higher \cite{liu2017delving,yuan2021meta,dong2018boosting}. 
Recently, some methods have combined ensemble-based transfer attacks with limited feedback from the victim models to improve the overall success rate \cite{guo2019simple,huang2019black,tashiro2020diversity,suya2019hybrid, li2021adversarial,lord2022attacking}. 
These methods have mainly focused on classification models, and ensemble attacks on dense prediction tasks such as object detection and semantic segmentation are relatively less studied, especially for targeted attacks \cite{wu2020making}.

\noindent \textbf{Attacks against object detectors and segmentation.}
Dense (pixel-level) prediction tasks such as object detection and semantic segmentation have higher task complexities\cite{wang2021pyramid} compared to classification tasks.  Existing attacks on object detectors mainly focus on whitebox setting, although there are a few exceptions \cite{wei2019transferable,cai2022blackbox}. A recent study \cite{cai2022context} generates blackbox attacks on object detectors by using a surrogate ensemble and context-aware attack-based queries. Another approach \cite{wei2019transferable} trains a generative model to generate transferable attacks. While some patch-based attacks \cite{Liu2019Dpatch,saha2020role} are effective, the patches are easily noticeable. 
Recent works \cite{gu2022segpgd,gu2021adversarial} have investigated adversarial robustness for semantic segmentation and proposed a transferable untargeted attack using a single surrogate model. 
While most existing methods are based on a single surrogate model, we demonstrate that using multiple surrogates with weight balancing/search in the attack generation process, we can generate more effective adversarial examples for both untargeted and targeted scenarios, as well as for various types of dense prediction tasks.

%% file: method.tex
\section{Method}

\subsection{Preliminaries}

We consider a per-instance attack scenario in which we generate adversarial perturbation $\delta$ for a given image $x$. To keep the perturbation imperceptible, we bound its $\ell_p$ norm as $\|\delta\|_p \leq \varepsilon$. In our experiments, we mainly use $\ell_\infty$ or max norm that limits the maximum level of perturbation. Our goal is to find $\delta$ such that the perturbed image, $x^\star = x + \delta$, can disrupt a victim image recognition system $f_\vv$ to make wrong predictions. Suppose the original prediction for the clean image $x$ is $y = f_\vv(x)$. The attack goal is $f(x^\star) \neq y$ for untargeted attack, and $f(x^\star) = y^\star$ for targeted attack, where $y^\star$ is the desired output (e.g., label or bounding box or segmentation map).

For classification models, the label $y \in \mathbb{R}$ is a scalar. However, dense prediction models can have  more complex output space. For object detection, the variable-length output $y \in \mathbb{R}^{K\times6}$, where $K$ is the number of detected objects, and each object label and position are encoded in a vector of length 6 that include the object category, bounding box coordinates, and confidence score. Some other tasks like keypoint detection and OCR are similar to object detection. For semantic segmentation, the prediction $y \in \mathbb{R}^{H \times W}$ is per-pixel classification, where $H$ and $W$ are the height and width of the input image, respectively. Depth and optical flow estimation tasks have similar output structure.

The adversarial loss functions for object detection and semantic segmentation can be defined using their respective training or prediction loss functions. 
Let us consider a whitebox model $f$ and an input image $x$ with output $y = f(x)$. For untargeted attack, we can search for the adversarial example $x^\star$ by solving the following maximization problem: 
\begin{equation}\label{eq:untargeted}
    x^\star = \argmax_{x} ~ \mathcal{L}(f(x), y),
\end{equation}
where $\mathcal{L}(f(x),y)$ represents the training loss of the model with input $x$ and output $y$. 
For targeted attacks, with a target output $y^\star$, we solve the following minimization problem: 
\begin{equation}\label{eq:targeted}
    x^\star = \argmin_{x} ~ \mathcal{L}(f(x), y^\star).
\end{equation}

Different from classification, which mostly use cross-entropy loss across different models, dense predictions have different loss functions for different models due to the complexity of the output space and diversity of the architectures. For example, two-stage object detector, including Faster RCNN \cite{ren2015faster}, has losses for object classification, bounding box regression, and losses on the region proposal network (RPN). But for one-stage object detectors like YOLO \cite{Redmon2016YOLO,redmon2018yolov3}, they do not have losses corresponding to RPN. Due to the large variability of the loss functions used in different dense prediction models, we use the corresponding training loss $\mathcal{L}$ for each model as the optimization loss to guide the backpropagation.

We employ PGD\cite{madry2017towards} to optimize the perturbation as 
\begin{equation}\label{eq:PGD}
    \delta^{t+1} = \Pi_{\varepsilon} \left( \delta^{t} - \lambda~ \mathbf{sign} (\nabla_{\delta}\mathcal{L}(f(x+\delta^t), y^\star)) \right),
\end{equation}
for targeted attack and
\begin{equation}\label{eq:PGD-untargeted}
    \delta^{t+1} = \Pi_{\varepsilon} \left( \delta^{t} + \lambda~ \mathbf{sign} (\nabla_{\delta}\mathcal{L}(f(x+\delta^t), y)) \right),
\end{equation}
for untargeted attack. Here $t$ indicates the attack step, $\lambda$ is the step size, and $\Pi_{\varepsilon}$ projects the perturbation into a $\ell_p$ norm ball with radius $\varepsilon$. In the rest of the paper,  we focus on targeted attacks without loss of generalization.

\subsection{Ensemble-based attacks}

In an ensemble-based transfer attack, we use an ensemble of $N$ surrogate (whitebox) models: $\mathcal{F} = \{f_1,\ldots, f_N\}$ to generate perturbations to attack the victim model $f_\vv$. Note that if the ensemble has a single model, then such an attack becomes a simple transfer attack with a single surrogate model. 
Let us denote the training loss function for $i$th model as $\mathcal{L}_i(f_i(x),y^*)$. A natural approach to combine the loss functions of all surrogate models is to compute an average or weighted average of the individual loss functions. For instance, we can generate the adversarial image by solving the following optimization problem: 
\begin{equation}\label{eq:ensemble-loss}
x^\star(\boldsymbol{\alpha}) = \argmin_{x} \sum_{i=1}^N \alpha_i\mathcal{L}_i(f_i(x),y^\star), 
\end{equation}
where $x^\star(\boldsymbol{\alpha})$ is a function of the weights of the ensemble ${\boldsymbol{\alpha} = \{\alpha_1,\ldots, \alpha_N\}}$. 
One of our key observations is that the choice of weights plays a critical role in the transfer attack success rate of the ensemble models.

\input{figures/fig-obj-loss-imbalance.tex}

\noindent\textbf{Weight balancing (victim model agnostic).} In ensemble-based transfer attacks, we build on the intuition that if an adversarial example can fool all models simultaneously, it would potentially be more transferable to any unseen victim model. This concept has been empirically corroborated by numerous works \cite{liu2017delving,dong2018boosting}. However, most attack methods have only been verified on classification models, all of which use the same cross-entropy loss and yield similar loss values. In contrast, the loss functions for object detectors in an ensemble can differ significantly and cover a large range of values (as shown in \cref{fig:obj-loss-imbalance}). In such cases, models with large loss terms heavily influence the optimization procedure, reducing the attack success rate for models with small losses (see~\cref{tab:obj-ablation}). To overcome this issue, we propose a simple yet effective solution to balance the weights assigned to each model in the ensemble as follows. 
For each input image $x$ and target output $y^\star$, we adjust the weight for $i$th surrogate model loss as 
\begin{equation}
\alpha_i = \frac{ \sum_{i=1}^N \mathcal{L}_i(f_i(x), y^\star)}{N \mathcal{L}_i(f_i( x), y^\star)}. \label{eq:weight-balancing}
\end{equation}
The weights are adjusted in a whitebox setting as it allows us to measure the loss of each whitebox model accurately. The purpose of weight balancing is to ensure that all surrogate models can be successfully attacked, making the generated example more adversarial for blackbox victim models.

\noindent\textbf{Weight optimization (victim model specific).} Note that the weight normalization, as discussed above, is agnostic to the victim model. We further observe that such transfer-based attacks can be further improved by optimizing the weights of the ensemble according to the victim model, input image, and target output. In particular, we can change the individual $\alpha_i$ to create the perturbations that reduce the victim model loss $\mathcal{L}_\vv$. To achieve this goal, we need to solve the following optimization problem with respect to $\boldsymbol{\alpha}$: 
\begin{equation}\label{eq:weight-optimization}
    \boldsymbol{\alpha}^\star = \argmin_{\boldsymbol{\alpha}} ~ \mathcal{L}_\vv(f_\vv(x^\star(\boldsymbol{\alpha})), y^\star).
\end{equation}
The optimization problem in \eqref{eq:weight-optimization} is a nested optimization that we can solve as an alternating minimization routine. \\
\textbf{Step 0.} Given input $x$, output $y^\star$, and surrogate ensemble $\mathcal{F}$, we initialize $\boldsymbol{\alpha}$ using \eqref{eq:weight-balancing}. \\
\textbf{Step 1.} Solve \eqref{eq:ensemble-loss} to generate an adversarial example $x^\star(\boldsymbol{\alpha})$. \\
\textbf{Step 2.} Test the victim model. Stop if attack is successful; otherwise, change one of the $\alpha_i$ and repeat Step 1. 

In our experiments, we update the $\alpha_i$ in a cyclic manner (one coordinate at a time) as $\alpha_i \pm \gamma$ in \textbf{Step 2}, where $\gamma$ denotes a step size. In every round, we select the value of $\alpha_i$ that provides smallest value of the victim loss. We count the number of queries as the number of times we test the generated adversarial example on the victim model and denote it as $Q$ in our experiments.

%% file: figures/fig-obj-loss-imbalance.tex
\begin{figure}[t]
    \centering
    \includegraphics[width=0.35\textwidth]{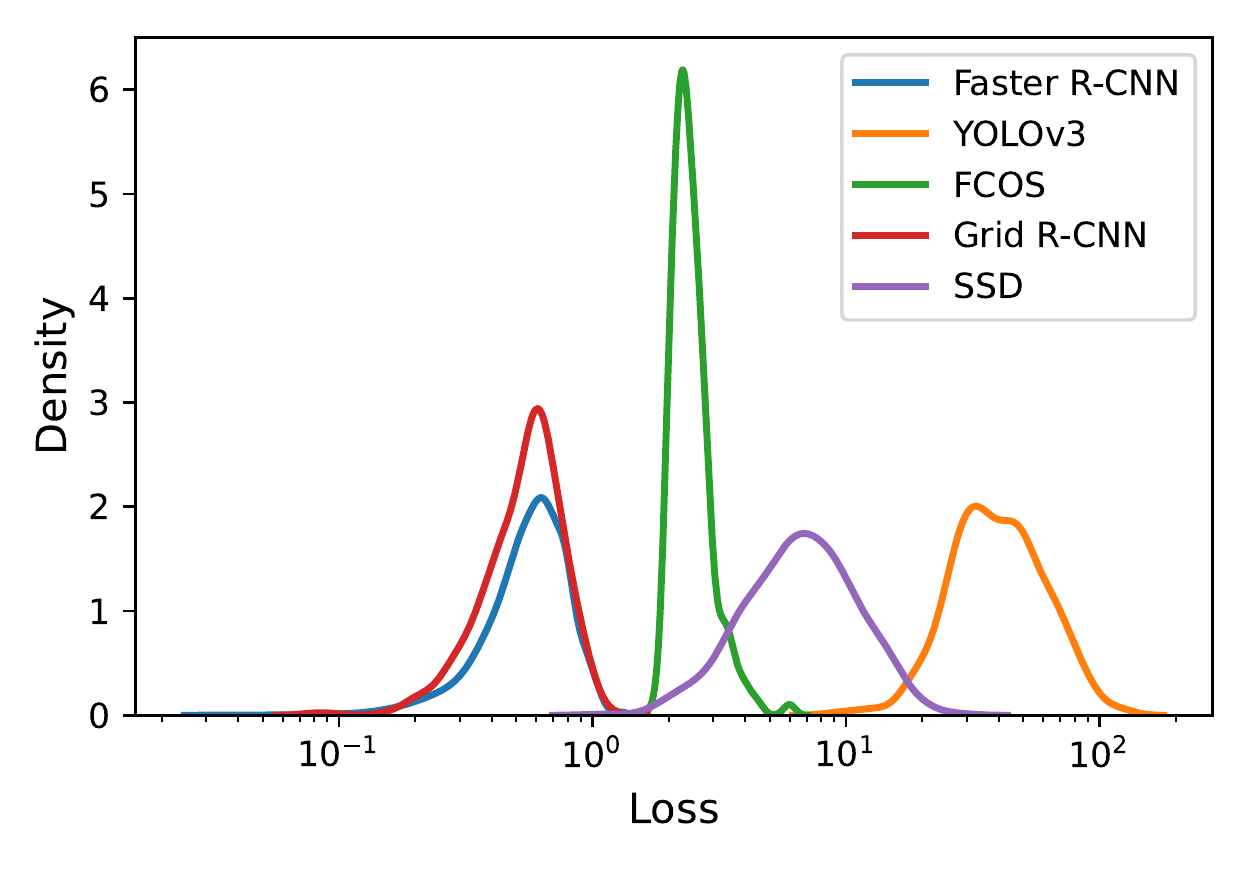}
    \vspace{-12pt}
    \caption{Distribution of losses for different object detection models. $\mathbb{P}(\mathcal{L}_i(f_i(x),y^\star)) $. Calculated on 500 images from VOC dataset.}
    \label{fig:obj-loss-imbalance}
    \vspace{-15pt}
\end{figure}

%% file: experiment.tex
\section{Experiments}

To evaluate the effectiveness of our method, we performed extensive experiments on attacking various object detection and semantic segmentation models. We first show that the attacks generated by a single surrogate model fail to transfer to arbitrary victim models. Then we show that the attack transfer rate can be increased by using an ensemble with weight balancing. Additional optimization of the weights surrogates for each victim model can further improve the attack performance. Finally, we show that we can generate single perturbations to fool object detectors and semantic segmentation models simultaneously.

\subsection{Experiment setup}

\subsubsection{Object detection}

\noindent\textbf{Models and datasets.} We utilize \texttt{MMDetection}~\cite{mmdetection} toolbox to select various model architectures and weights pre-trained on COCO 2017 dataset \cite{lin2014microsoft}. To construct the surrogate ensemble, we start with two widely used models, \texttt{Faster R-CNN}~\cite{ren2015faster} and \texttt{YOLO}~\cite{Redmon2016YOLO,redmon2018yolov3}, and expand the ensemble by appending models with different architectures, including \{\texttt{FCOS}~\cite{tian2019fcos}, \texttt{Grid R-CNN}~\cite{lu2019grid}, \texttt{SSD}~\cite{liu2016ssd}\}. 
 We select different victim models, including \{\texttt{RetinaNet}~\cite{lin2017focal}, \texttt{Libra R-CNN}~\cite{pang2019libra}, \texttt{FoveaBox}~\cite{kong2020foveabox}, \texttt{FreeAnchor}~\cite{zhang2019freeanchor}, \texttt{DETR}~\cite{carion2020end}\}. 
We evaluate attack performance on COCO 2017 \cite{lin2014microsoft} and Pascal VOC 2007 \cite{everingham2010pascal} datasets. Since the models from this repository are trained on COCO, which contains 80 object categories (a superset of VOC dataset's 20 categories), while testing on VOC dataset, we only return the objects that exist in VOC. We follow the setup in \cite{cai2022context} and randomly select 500 images containing multiple (2--6) objects from VOC 2007 test and COCO 2017 validation sets.

\noindent\textbf{Evaluation metrics.} We mainly focus on targeted attacks for object detection since they are more challenging than untargeted or vanishing attacks. We measure the performance of the attack using attack success rate (ASR), which equals the number of successfully attacked images over the total number of attacks. We follow the setting in \cite{cai2022context}, where if the target label is detected within the victim object region with IOU $> 0.3$, the attack is determined a success.

\noindent\textbf{Perturbation and query budget.} We tested different perturbation levels with $\ell_\infty = \{ 10,20,30\}$ out of $255$. We use at most 10 queries for attacking object detectors, and we show the trends of how ASR increases with the number of queries. To align with \cite{cai2022context} that uses 5 attack plans, we set the maximum query budget to $Q = 5$ in \cref{tab:obj-ablation}.

\noindent\textbf{Comparing methods.} 
We compare with \cite{cai2022context}, which is a state-of-the-art transfer-based approach that leverages context information to design attack plans to iteratively attack the victim object. The method generates different perturbations by iterating over a set of predefined attacks, and the total number of queries is the number of attempted attacks. BASES \cite{cai2022blackbox} is a recent work on ensemble-based blackbox attacks, which mainly focused on classification tasks and did not consider the loss distributions of different surrogate models. In our experiments, the ensemble with weight optimization and without balancing is equivalent to BASES \cite{cai2022blackbox}.

\input{tables/tab-obj-ablation.tex}
\input{figures/fig-obj-trend.tex}

\subsubsection{Semantic segmentation}

\noindent\textbf{Models and datasets.}
We use \texttt{MMSegmentation}~\cite{mmseg2020} toolbox to select different model architectures and weights pre-trained on Cityscapes \cite{cordts2016cityscapes} ($x \in \mathbb{R}^{512\times1024\times3}$)  and Pascal VOC ($x \in \mathbb{R}^{512\times512\times3}$) datasets.
We select \texttt{PSPNet}~\cite{zhao2017pyramid} and \texttt{DeepLabV3}~\cite{chen2017rethinking} with \texttt{ResNet50} and \texttt{ResNet101} \cite{he2016deep} backbones as our blackbox victim models.
For the surrogate ensemble, we start with the primary semantic segmentation model \texttt{FCN}~\cite{long2015fcn}, and expand the ensemble 
with \{\texttt{UPerNet}~\cite{xiao2018upernet}, \texttt{PSANet}~\cite{zhao2018psanet}, \texttt{GCNet}~\cite{cao2019gcnet}, \texttt{ANN}~\cite{zhu2019ann}, \texttt{EncNet}~\cite{Zhang_2018_encnet}\}. All models are built on \texttt{ResNet50}~\cite{he2016deep} backbone trained with the cross-entropy loss. 
The loss values across all surrogate models have similar range; therefore, the effect of weight balancing for semantic segmentation is not as significant as it is for object detection.  
We use validation datasets from Cityscapes \cite{cordts2016cityscapes} and Pascal VOC 2012, which contains 500 and 1499 images with 19 and 21 classes,  respectively.

\noindent\textbf{Evaluation metrics.}
We use different metrics for untargeted and targeted attack performance evaluation. In untargeted experiments, the attack performance is evaluated using the mIoU score (in percentage \%), the lower mIoU score the better attack performance. For targeted experiments, we report the pixel success ratio (PSR), which indicates the percentage of pixels successfully assigned the desired label in the target region, the higher the better attack performance. %

\noindent\textbf{Perturbation and query budget.}
We use the perturbation budget $\ell_\infty \leq 8$ out of 255 and 
query budget $Q=20$.

\noindent\textbf{Comparing methods.} 
We compare with dynamic scale (DS) attack \cite{gu2021adversarial} which is the most recent method that achieves the highest attack transfer rate on semantic segmentation untargeted attacks.

\subsection{Attacks against object detection}
Following settings in \cite{cai2022context}, we randomly select one object from the output of victim model as the victim object and perturb it into a target object that does not exist in the original detection. This approach rules out the possibility of mis-counting existing objects as the target object. 

\input{tables/tab-obj-ensemble-size.tex}

\input{tables/tab-seg-untar-cs.tex}
\input{tables/tab-seg-tar.tex}

We report our main results in \cref{tab:obj-ablation}. 
The baseline method uses a surrogate ensemble without weight balancing and models are assigned weight of 1. 
Such a baseline method is same a transfer-based method and results in highly imbalanced success rate for different surrogate models. For instance, at $\ell_\infty \leq 10$, the success rate for \texttt{YOLOv3} is above 90\% while the success rate for \texttt{Faster R-CNN} is less than 30\%. Low success rate on surrogate side translates to low success rate on blackbox victim side. The main reason for such imbalance is that the loss of different object detectors can be highly unbalanced (e.g., the loss value for \texttt{YOLOv3} is nearly 60$\times$ larger than the loss of \texttt{Faster RCNN} for targeted attacks, \cf \cref{fig:obj-loss-imbalance}). 
With weight balancing, the success rate increases for surrogate and blackbox victim models. The success rate is further increased on surrogate and victim blackbox models if we optimize the weights, same as BASES \cite{cai2022blackbox}.
Our method (with weight balancing and optimization) achieves a significantly higher ASR compared to context-aware attack across different datasets and different perturbation budgets. On average, our ASR on blackbox victim models is over $4\times$ better than baseline method and over $1.5\times$ better than context-aware attack. On whitebox surrogate models, weight balancing and optimization also achieves the highest ASR. Context-aware attack fixes weight ratio for surrogate models, $\nicefrac{\alpha_{\text{FRCNN}}}{\alpha_{\text{YOLO}}}  = 4$, which is sub-optimal according to our analyses. Even though it achieves much higher performance than baseline, it still largely under-performs our method. 
Similar trend is observed for COCO dataset (see~\cref{tab:obj-ablation-coco}).

\cref{fig:obj-trend} shows the effect of the number of queries on the ASR that gradually improves as we optimize the weights. We observe the largest increase in the first two steps and then the improvement plateaus as $Q\rightarrow10$.

We also conducted an experiment to test our method with varying ensemble sizes. The results for $\ell_\infty \le 10$, $Q=5$ are presented in \cref{tab:obj-ensemble-size}. As we increase the number of models in the ensemble from $N=1$ to $N=5$, we observe an increased ASR on all blackbox victim models.

\subsection{Attacks against semantic segmentation}
We evaluate the effectiveness of our attack on semantic segmentation in both untargeted and targeted settings. 
For the sake of consistency and a fair comparison, we adopt adversarial attack settings in DS attack \cite{gu2021adversarial}. 

\input{figures/fig-seg-trend.tex}

\noindent \textbf{Untargeted attacks.}
We generate adversarial attacks using different ensemble sizes and report mIoU scores on Cityscapes in \cref{tab:seg-untar-cs} and \cref{fig:seg-trend} (and Pascal VOC in supplementary material).
In the untargeted setting, semantic segmentation models are attacked to maximize the loss between clean and modified annotation; hence, the lower mIoU implies better attack performance. All of the victim models achieve high performance on clean images. 
The baseline method (direct transfer attack with one surrogate model using PGD) performs well in the whitebox setting but suffers when the victim uses another backbone. For example, the attacks generated on \texttt{PSPNet-Res50} achieves 3.43\% mIoU on \texttt{PSPNet-Res50} but only attains 24.18\% mIoU on \texttt{PSPNet-Res101}. DS attack achieves better results than the baseline method but still suffers from cross-backbone transfers. 
On the other hand, our method, without weight optimization (\ie, $Q=0$) and using a surrogate ensemble of $N=2$ models, can achieve results comparable to DS attack, particularly for attacks on Res101 models. As we increase the number of surrogate models to 4 or 6, our attack performance further improves. 
Furthermore, when we apply weight optimization (\eg, $Q=20$), the attack improves by updating the weights of the surrogate models, allowing us to outperform DS attack for all victim models. 
\cref{fig:seg-trend} shows how the mIoU changes with the number of queries. We observe that the mIoU gradually reduces as we query the victim model and optimize the weights. The largest decrease happens in the first 3--4 steps and then the reduction plateaus as $Q\rightarrow20$.

\input{figures/fig-joint-curves-supp.tex}

\input{figures/fig-joint-attack.tex}

\noindent \textbf{Targeted attack.}
To evaluate our method in a more challenging setting, we consider a targeted attack scenario, where instead of changing every pixel in the segmentation to some arbitrary label, we focus on attacking a dominant class (\ie, the class occupying the largest area) in the scene to its least likely class $y^\star$. For each clean image, we first select a region with the dominant class $y$ (\eg, ``road'' or ``building'' for most of the Cityscapes images. See~\cref{fig:seg-tgt-explain} as an example). Then based on the least-likely class of each pixel in that region,
we select the class that appears most frequently as the target label $y^\star$ of the entire region.
We use PSR as our evaluation metric, which represents the percentage of pixels in the selected region that are successfully assigned to $y^\star$. The higher percentage indicates more pixels are successfully attacked to the desired class, which indicates better attack performance. %
Our targeted attack results are reported in \cref{tab:seg-targeted}.
Results show that as we increase the number of surrogate models $(N)$, the ASR improves for most instances without any weight optimization step (i.e., $Q=0$). If we perform weight optimization for $Q=20$ steps, then the success rate increases for all the models. For instance, with $N=4$, the ASR for Res101 models increases from 17--26\% to 63-64\%.

\subsection{Joint attack for multiple models and tasks}
We first show that generally adversarial examples generated for object detection do not transfer to semantic segmentation, and vice versa. Then we show that we can generate single perturbations to fool object detectors and semantic segmentation models simultaneously, by using a surrogate ensemble including both detection and segmentation models. We choose targeted attacks in our experiments because they are more challenging than untargeted attacks. 

\noindent \textbf{Experiment setup.} On the blackbox (victim) side, we tested \texttt{RetinaNet} as the victim object detector and \texttt{PSPNet-Res50} as the victim semantic segmentation model. On the whitebox (surrogate) side, we used \texttt{Faster RCNN, YOLOv3} as the surrogate object detectors and \texttt{FCN, UPerNet} as the surrogate semantic segmentation models. We performed targeted attacks on 500 test images selected from the validation set of CityScapes dataset.

\noindent \textbf{Results.} We present the ASRs for task-specific and joint attacks in \cref{fig:obj-joint-curves}. Green curves denote ASR for object detectors, and blue curves denote PSR for semantic segmentation. 
\cref{fig:sub-joint-det} presents the results when we generate attacks using an object detector surrogate ensemble. Note that success rate for victim object detector (\texttt{RetinaNet}) increases as we optimize the weights but the  success rate for the semantic segmentation model (\texttt{PSPNet}) remains small. 
Similarly, \cref{fig:sub-joint-seg} presents the results when we generate attacks using a segmentation surrogate ensemble. The success rate for the victim semantic segmentation model increases, but the success rate for the object dector remains close to zero. 
\cref{fig:sub-joint-all} presents the results when we perform a joint attack using an ensemble that consists of both object detectors and segmentation models. The blackbox ASR is high on both detection and segmentation \cref{fig:sub-joint-all}, and the attack performance improves as we update the weights of the surrogate models. In \cref{fig:sub-joint-all}, we show the results for different perturbation budgets, with $\ell_\infty \leq 10$, the success rates on detection and segmentation are between $60\%-70\%$, which are close to in-domain detection attacks in \cref{fig:sub-joint-det} and in-domain segmentation attacks \cref{fig:sub-joint-seg}. When we increase the perturbation to $\ell_\infty \leq 20$, the success rate for both detection and segmentation can surpass $80\%$.

\noindent \textbf{Visualization of adversarial examples.} 
In this example, our goal is to perturb the car in the middle to a traffic light. We assign the target label for car region to traffic light. \cref{fig:obj-joint-attack} shows the results where a single adversarial image generated by the surrogate model can successfully fool the blackbox models \texttt{RetinaNet} and \texttt{PSPNet}. 

%% file: tables/tab-obj-ablation.tex
\begin{table*}[]

\centering
\small
\caption{Targeted attack success rate (\%) of different methods at different perturbation budgets on VOC dataset. For each perturbation level, the first 4 rows correspond to different settings of our attacks, \ie with (\cmark) or without (\xmark) weight balancing and weight optimization. We show comparison with context-aware attack \cite{cai2022context}, the state-of-the-art method for query-based blackbox attacks.}
\label{tab:obj-ablation}

\begin{tabular}{ccccc|ccccc}
\hline
\multirow{2}{*}{\begin{tabular}[c]{@{}c@{}}Perturbation \\ Budget\end{tabular}} & \multirow{2}{*}{\begin{tabular}[c]{@{}c@{}}Weight \\ Balancing\end{tabular}} & \multirow{2}{*}{\begin{tabular}[c]{@{}c@{}}Weight \\ Optimization\end{tabular}} & \multicolumn{2}{c|}{\textbf{Surrogate Ensemble}} & \multicolumn{5}{c}{\textbf{Blackbox Victim Models} (ASR $\uparrow$)} \\
 &  &  & FRCNN & YOLOv3 & Retina & Libra & Fovea & Free & DETR \\ \hline
\multirow{5}{*}{$\ell_{\infty} = 10$} & \xmark & \xmark & 27.9 & 91.5 & 11.6 & 9.2 & 9.0 & 13.4 & 5.6 \\
 & \xmark & \cmark & 61.4 & \textbf{99.4} & 24.3 & 28.0 & 22.4 & 31.0 & 15.4 \\
 & \cmark & \xmark & 71.1 & 85.7 & 30.9 & 33.4 & 27.2 & 36.0 & 12.2 \\
 & \cmark & \cmark & \textbf{86.0} & 96.9 & \textbf{53.2} & \textbf{56.6} & \textbf{47.2} & \textbf{57.4} & \textbf{29.0} \\
 & \multicolumn{2}{c}{Context-aware Attack \cite{cai2022context}} & 55.8 & 75.6 & 22.6 & 20.4 & 33.6 & 39.2 & 20.2 \\ \hline
\multirow{5}{*}{$\ell_{\infty} = 20$} & \xmark & \xmark & 40.1 & 92.2 & 16.9 & 20.4 & 15.4 & 23.2 & 9.7 \\
 & \xmark & \cmark & 77.7 & \textbf{99.8} & 41.0 & 45.4 & 37.8 & 47.0 & 22.5 \\
 & \cmark & \xmark & 82.7 & 89.8 & 41.0 & 50.4 & 44.8 & 57.0 & 21.6 \\
 & \cmark & \cmark & \textbf{94.6} & 98.0 & \textbf{66.9} & \textbf{74.4} & \textbf{68.0} & \textbf{79.4} & \textbf{48.0} \\
 & \multicolumn{2}{c}{Context-aware Attack \cite{cai2022context}} & 78.6 & 87.2 & 35.2 & 38.4 & 51.6 & 56.6 & 34.0 \\ \hline
\multirow{5}{*}{$\ell_{\infty} = 30$} & \xmark & \xmark & 43.4 & 91.1 & 17.1 & 22.6 & 17.4 & 27.2 & 11.4 \\
 & \xmark & \cmark & 82.7 & \textbf{99.6} & 47.2 & 54.8 & 47.0 & 57.4 & 33.4 \\
 & \cmark & \xmark & 85.3 & 90.2 & 48.8 & 56.8 & 45.6 & 59.6 & 29.2 \\
 & \cmark & \cmark & \textbf{96.0} & 98.1 & \textbf{78.9} & \textbf{82.8} & \textbf{76.8} & \textbf{83.0} & \textbf{58.8} \\
 & \multicolumn{2}{c}{Context-aware Attack \cite{cai2022context}} & 80.6 & 88.0 & 42.0 & 44.2 & 56.8 & 63.6 & 40.2 \\ \hline
\end{tabular}

\end{table*}

%% file: figures/fig-obj-trend.tex
\begin{figure*}[!ht]
    \centering 
    \begin{subfigure}[t]{0.3\linewidth}
    \includegraphics[width=1\linewidth]{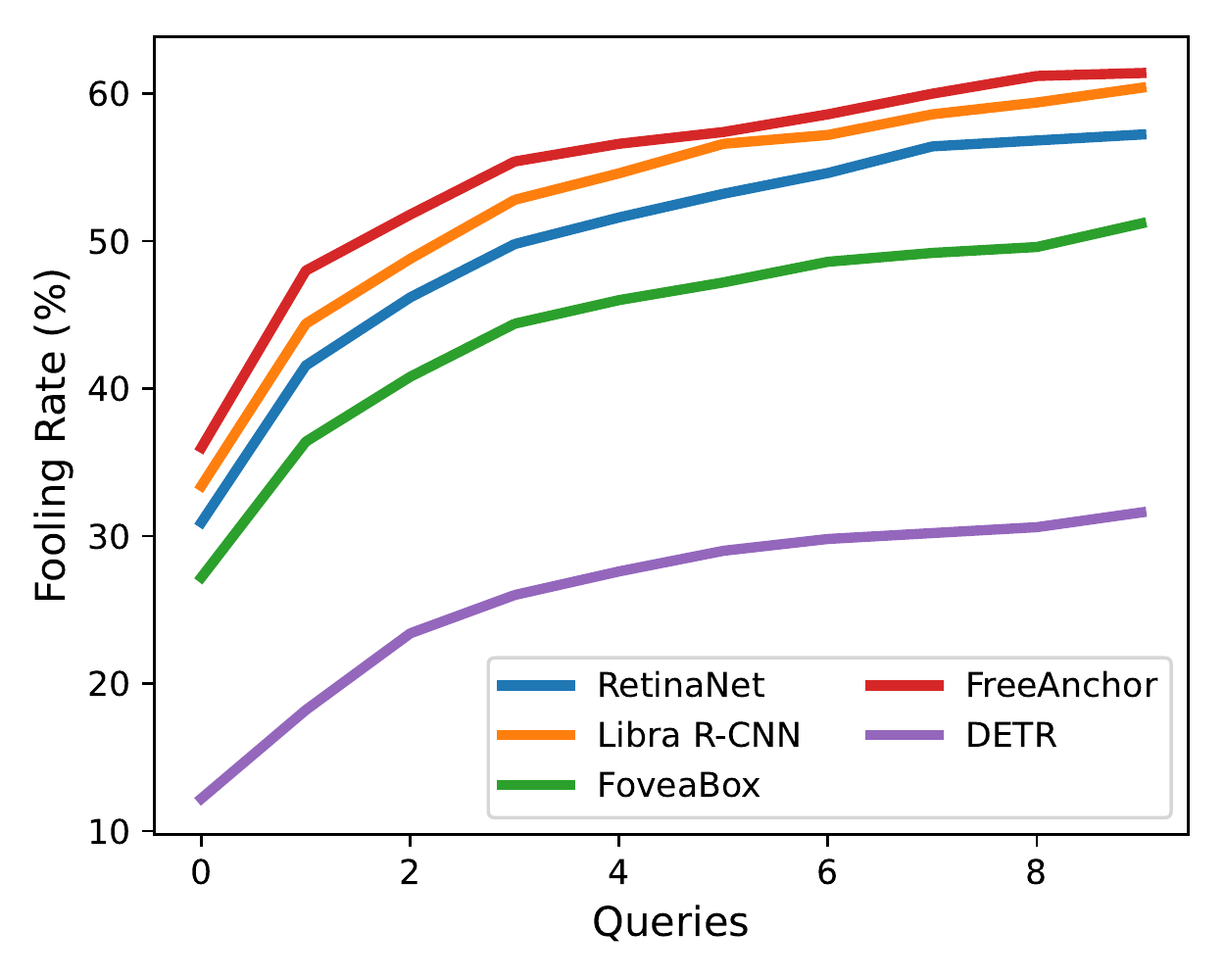}
    \caption{$\ell_{\infty} \leq 10$}
    \end{subfigure}
    ~~
    \begin{subfigure}[t]{0.3\linewidth}
    \includegraphics[width=1\textwidth]{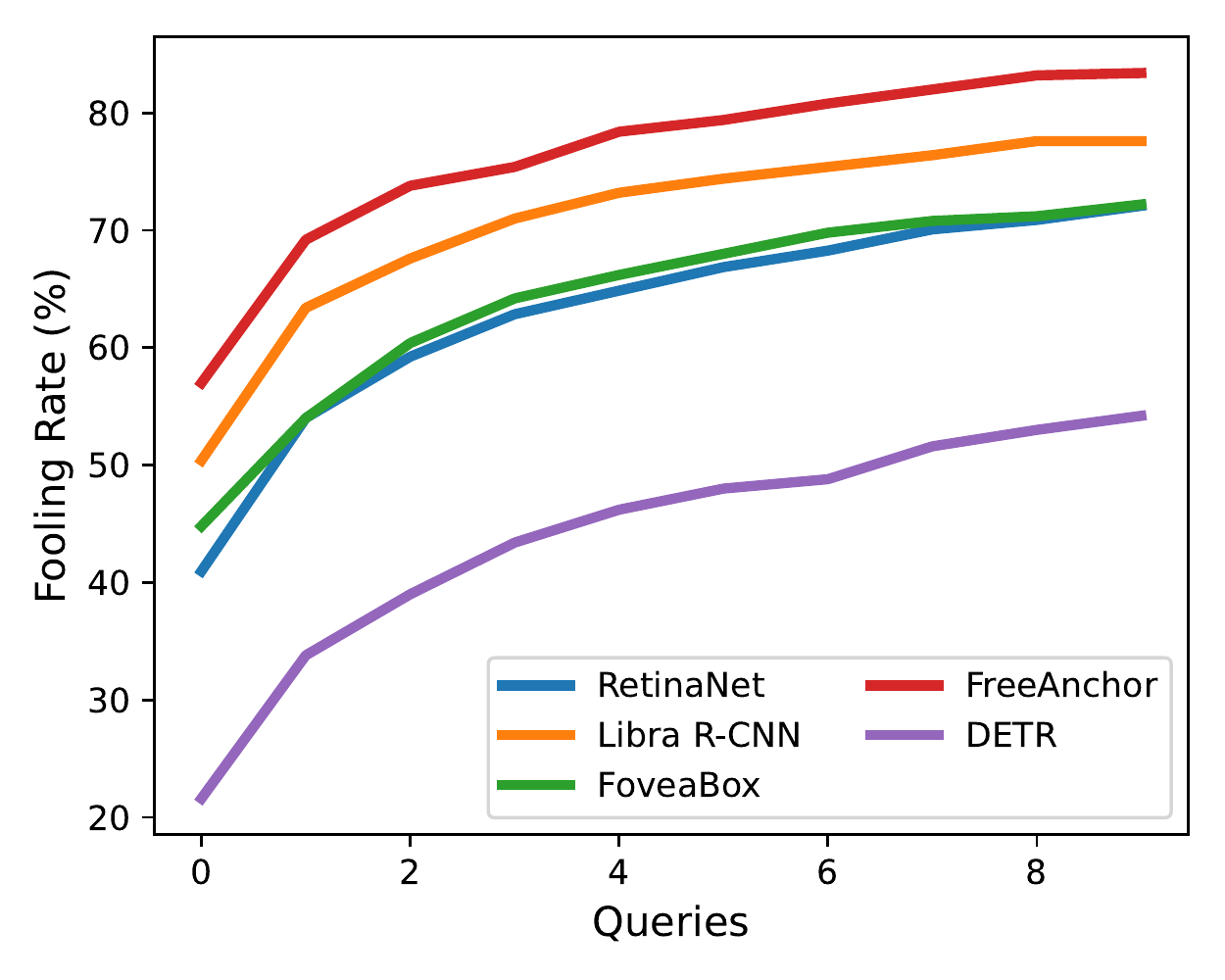}
    \caption{$\ell_{\infty} \leq 20$}
    \end{subfigure}
    ~~
    \begin{subfigure}[t]{0.3\linewidth}
    \includegraphics[width=1\textwidth]{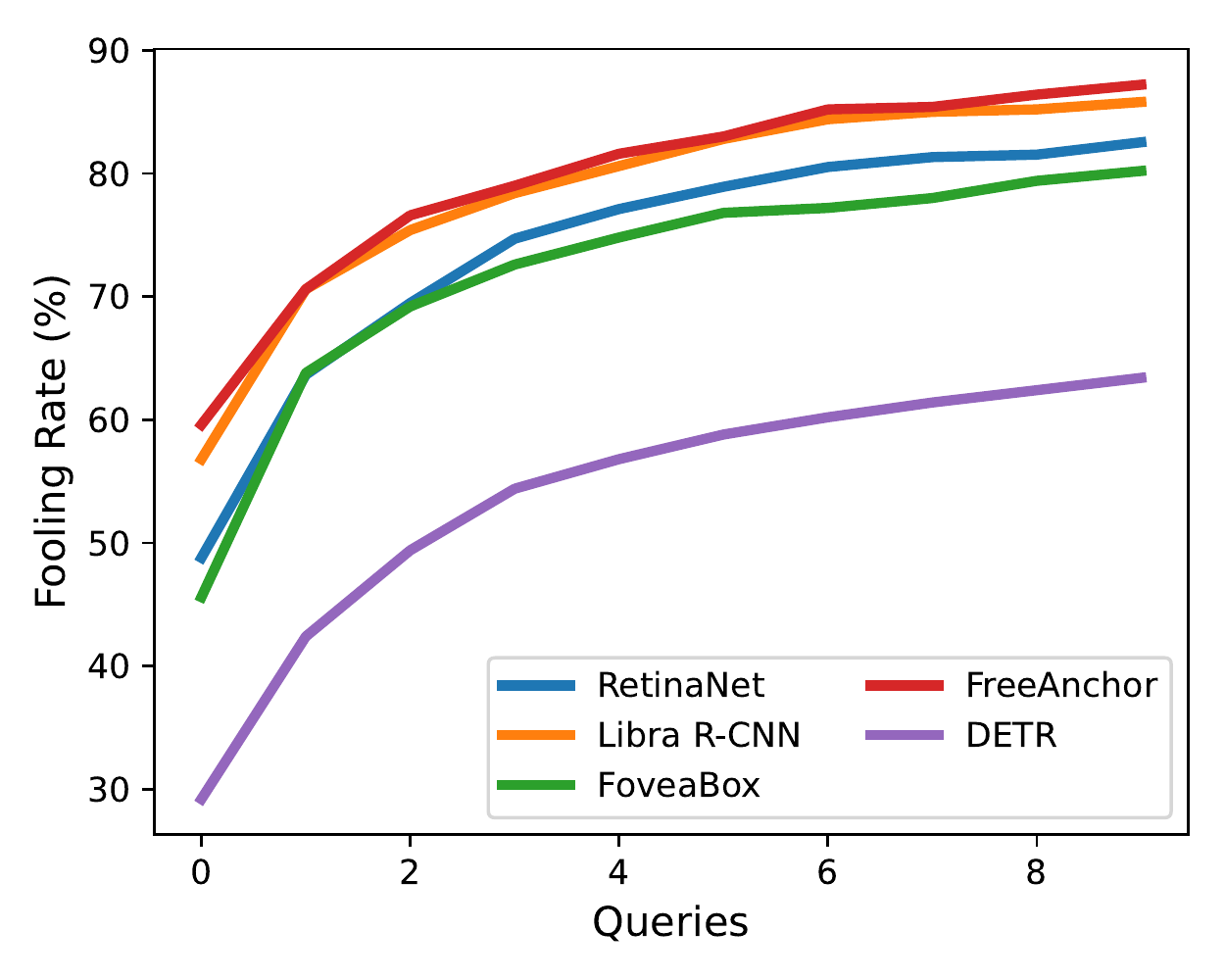}
    \caption{$\ell_{\infty} \leq 30$}
    \end{subfigure} 
    \caption{Attack success rate (or fooling rate) vs number of queries $(Q)$. The maximum value of $Q$ is set to 10 for these results.
    }
    \label{fig:obj-trend}
\end{figure*}

%% file: tables/tab-obj-ensemble-size.tex
\begin{table*}[]

\centering
\small
\caption{Targeted ASR (\%) for blackbox victim models and whitebox surrogate models with different ensemble sizes ($N$). On VOC dataset, $\ell_\infty \leq 10$.}
\label{tab:obj-ensemble-size}

\begin{tabular}{cccccc|ccccc}
\hline
\multirow{2}{*}{$N$} & \multicolumn{5}{c|}{\textbf{Surrogate Ensemble}} & \multicolumn{5}{c}{\textbf{Blackbox Victim Models} (ASR $\uparrow$)} \\
 & FRCNN & YOLOv3 & FCOS & Grid R-CNN & SSD & Retina & Libra & Fovea & Free & DETR \\ \hline
1 & 74.7 & - & - & - & - & 31.3 & 31.2 & 29.8 & 40.4 & 10.6 \\
2 & 86.0 & 96.9 & - & - & - & 53.2 & 56.6 & 47.2 & 57.4 & 29.0 \\
3 & 87.9 & 96.1 & 74.2 & - & - & 63.1 & 62.0 & 57.3 & 66.6 & 38.0 \\
4 & 89.6 & 94.7 & 75.2 & 87.9 & - & 68.7 & \textbf{71.0} & 67.6 & 74.4 & 49.6 \\
5 & 89.7 & 91.8 & 73.5 & 86.1 & 82.4 & \textbf{68.9} & 70.2 & \textbf{68.4} & \textbf{77.6} & \textbf{53.2} \\ \hline
\end{tabular}
\end{table*}

%% file: tables/tab-seg-untar-cs.tex
\begin{table*}[h]
\centering
\small
\caption{Untargeted attack mIoU scores (\%) of ensemble sizes $N=2,4,6$ on Cityscapes dataset. We compare $Q=0$ (i.e. direct transfer attack) with $Q=20$ ensemble attack performance. DS uses DeepLabV3-Res50 (DL3-50) as the surrogate model for attack generation; thus the DS on DL3-50 is a whitebox attack. While our method used an ensemble that does not include any victim models for attack generation, we still achieved comparable mIoU scores to DS on DL3-50. \textcolor{blue}{Blue} numbers represent whitebox attacks.}
\label{tab:seg-untar-cs}

\begin{tabular}{c|c|cccc}
\hline
 &  & \multicolumn{4}{c}{\textbf{Blackbox Victim Models} (mIoU $\downarrow$)}\\
\multirow{-2}{*}{Method} & \multirow{-2}{*}{\textbf{Whitebox Surrogate}} & PSPNet-Res50 & PSPNet-Res101 & DeepLabV3-Res50 & DeepLabV3-Res101 \\ \hline
Clean Images & - & 77.92 & 78.28 & 79.12 & 77.12 \\ \hline
 & PSPNet-Res50 & \textcolor{blue}{3.43} & 24.18 & 5.05 & 25.74 \\
\multirow{-2}{*}{Baseline} & DeepLabV3-Res50 & 4.76 & 21.72 & \textcolor{blue}{3.92} & 22.23 \\ \hline
\cellcolor[HTML]{FFFFFF} & PSPNet-Res50 & \textcolor{blue}{0.82} & 8.04 & 1.36 & 9.00 \\
\multirow{-2}{*}{\cellcolor[HTML]{FFFFFF}DS\cite{gu2021adversarial}} & DeepLabV3-Res50 & 1.23 & 7.97 & \textcolor{blue}{0.61} & 7.11 \\ \hline
 & $N=2$ & 5.07 & 8.32 & 5.19 & 8.74 \\
 & $N=4$ & 4.33 & 6.26 & 4.32 & 6.33 \\
\multirow{-3}{*}{Ours ($Q = 0$)} & $N=6$ & 3.62 & 4.91 & 4.02 & 4.84 \\ \hline
 & $N=2$ & 1.38 & 2.88 & 1.15 & 3.50 \\
 & $N=4$ & \textbf{0.79} & 2.04 & \textbf{0.73} & 1.80 \\
\multirow{-3}{*}{Ours ($Q = {20}$)} & $N=6$ & 0.90 & \textbf{1.55} & 0.94 & \textbf{1.09} \\ \hline
\end{tabular}

\end{table*}

%% file: tables/tab-seg-tar.tex
\begin{table} 
\centering
\small
\caption{
Targeted attack performance on Cityscapes as pixel success rate (higher the better).  The attack performance increases as we increase ensemble size ($N$) and number of queries for weight optimization ($Q$). $N=1$ has zero query.  We note PSPNet-Res50 as PSP-r50, and DeepLabV3-Res50 as DL3-r50, similar abbreviations apply to Res101.}
\vspace{5pt}
\label{tab:seg-targeted}
\begin{tabular}{c|c|cccc}
\hline
 &  & \multicolumn{4}{c}{\textbf{Blackbox Victim Models} (PSR $\uparrow$)} \\
\multirow{-2}{*}{$Q$} & \multirow{-2}{*}{$N$} & PSP-r50 & PSP-r101 & DL3-r50 & DL3-r101 \\ \hline
 & 1 & 39.15 & 10.21 & 35.02 & 7.58 \\
 & 2 & 52.15 & 12.28 & 47.99 & 10.59 \\
 & \cellcolor[HTML]{FFFFFF}3 & 43.17 & 11.34 & 42.10 & 9.87 \\
 & \cellcolor[HTML]{FFFFFF}4 & 51.44 & 26.13 & 49.14 & 17.42 \\
\multirow{-5}{*}{\begin{tabular}[c]{@{}c@{}} 0\end{tabular}} & \cellcolor[HTML]{FFFFFF}5 & 52.24 & 23.88 & 51.75 & 16.08 \\ \hline
\cellcolor[HTML]{FFFFFF} & \cellcolor[HTML]{FFFFFF}2 & 83.97 & 51.80 & 82.70 & 46.95 \\
\cellcolor[HTML]{FFFFFF} & \cellcolor[HTML]{FFFFFF}3 & 88.88 & 64.63 & 85.55 & 60.88 \\
\cellcolor[HTML]{FFFFFF} & \cellcolor[HTML]{FFFFFF}4 & 91.51 & 64.28 & 87.19 & 63.88 \\
\multirow{-5}{*}{\cellcolor[HTML]{FFFFFF}\begin{tabular}[c]{@{}c@{}} {20}\end{tabular}} & \cellcolor[HTML]{FFFFFF}5 & \textbf{92.91} & \textbf{69.09} & \textbf{88.95} & \textbf{69.65} \\ \hline
\end{tabular}
\end{table}

%% file: figures/fig-seg-trend.tex
\begin{figure*}[h]
    \centering 
    \begin{subfigure}[t]{0.3\linewidth}
    \includegraphics[width=1\linewidth]{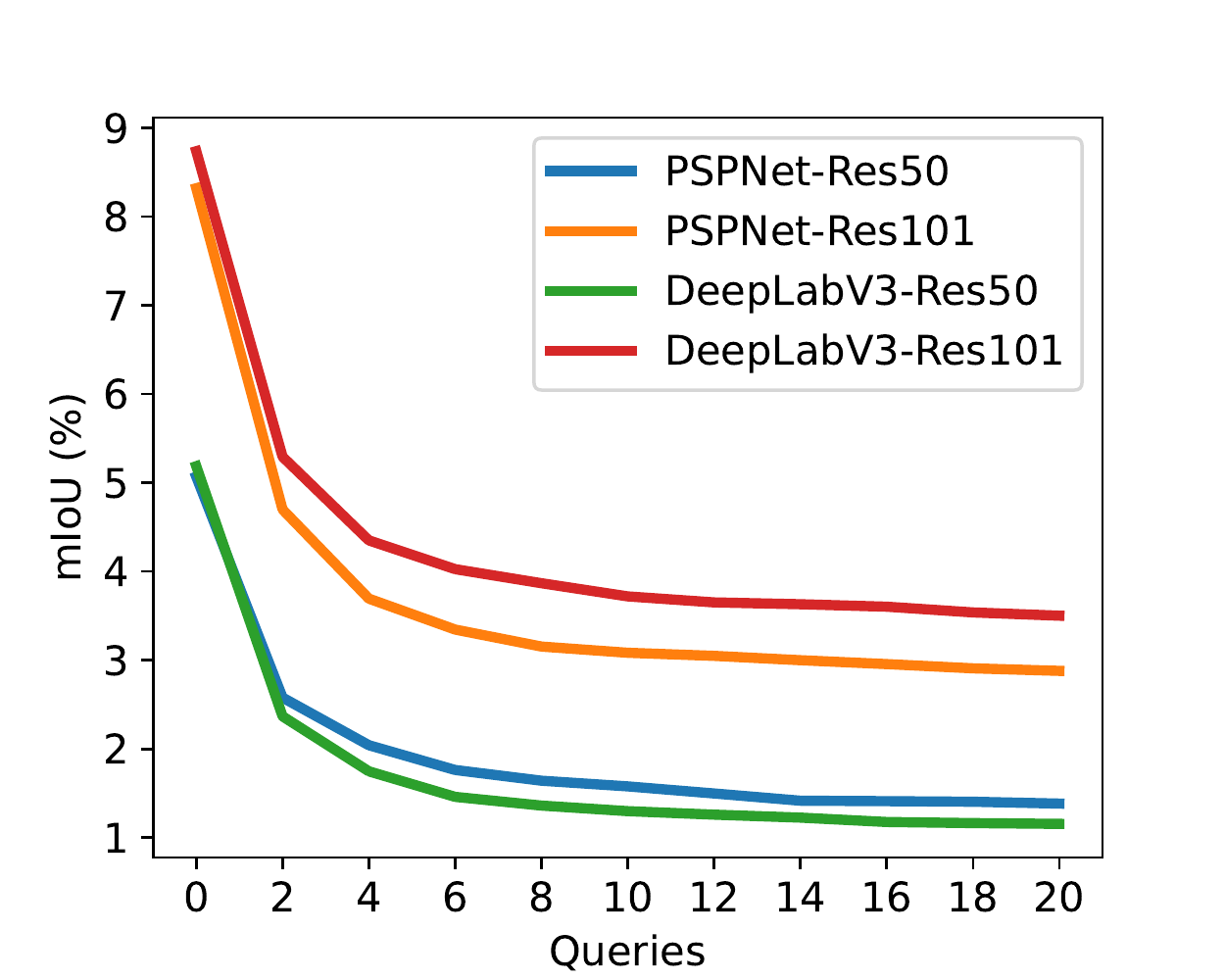}
    \caption{$N= 2$}
    \end{subfigure}
    ~~
    \begin{subfigure}[t]{0.3\linewidth}
    \includegraphics[width=1\textwidth]{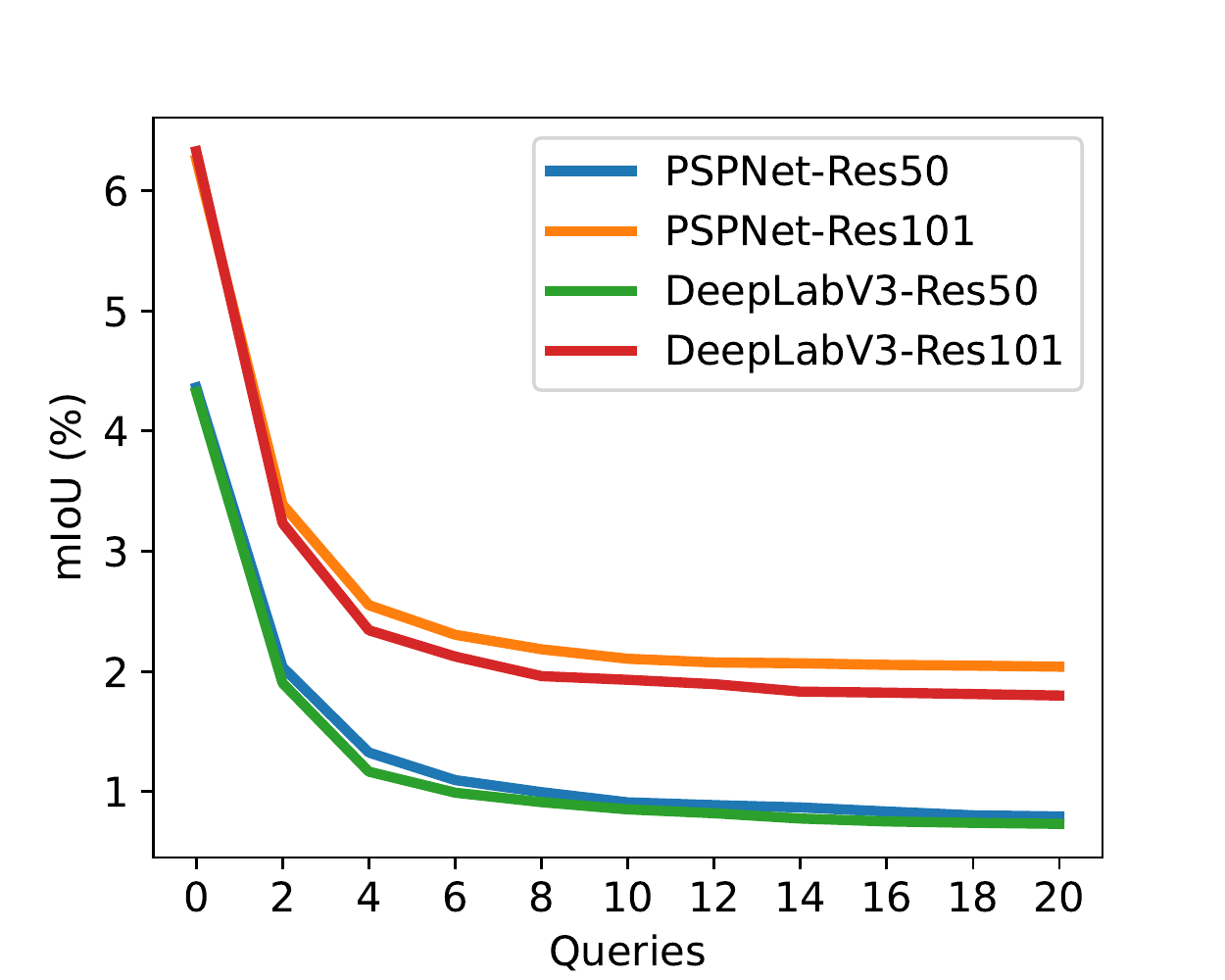}
    \caption{$N= 4$}
    \end{subfigure}
    ~~
    \begin{subfigure}[t]{0.3\linewidth}
    \includegraphics[width=1\textwidth]{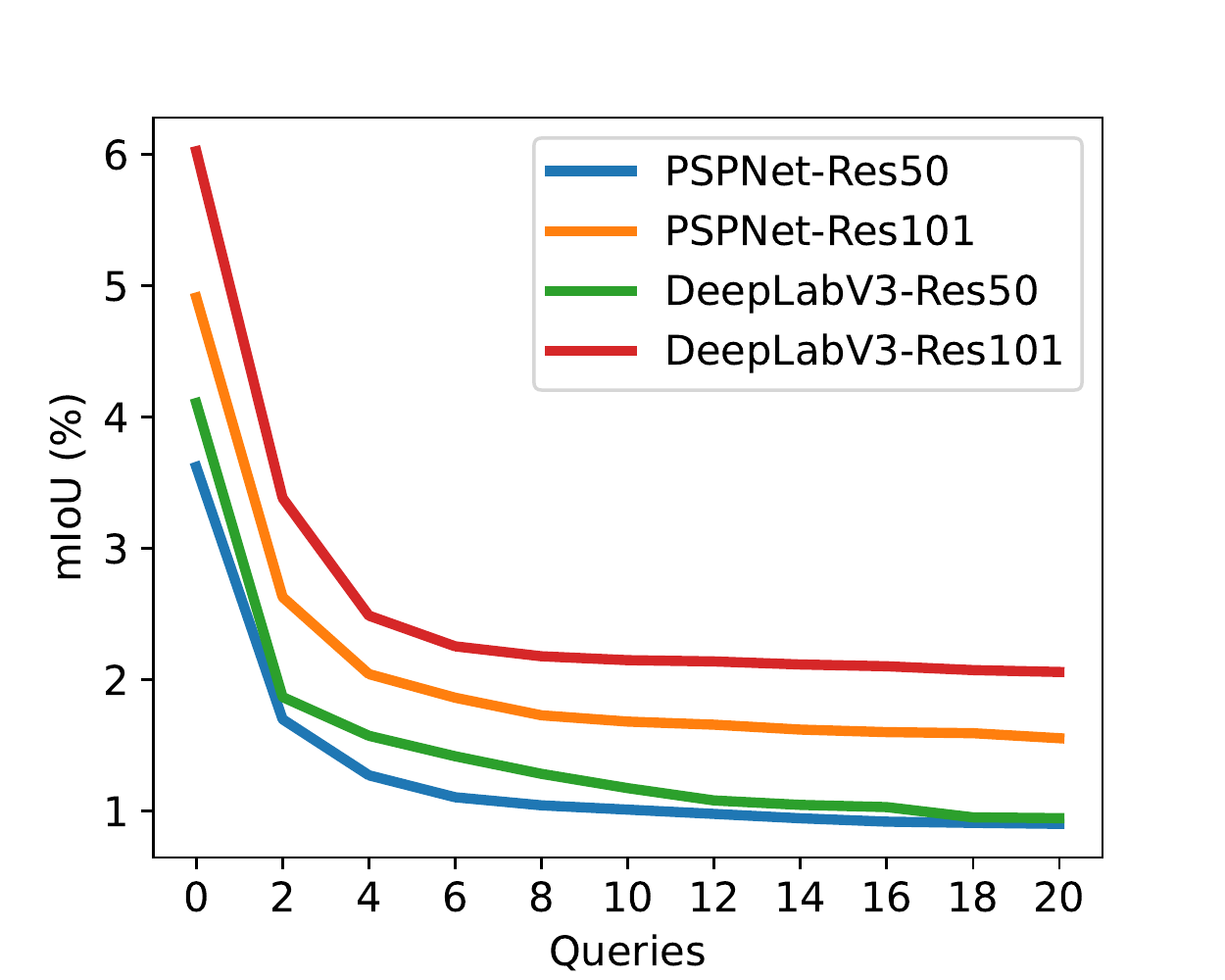}
    \caption{$N = 6$}
    \end{subfigure}
    \caption{mIoU vs number of queries $(Q)$ for different ensemble sizes $(N)$. 
    }
    \label{fig:seg-trend}
    \vspace{-10pt}
\end{figure*}

%% file: figures/fig-joint-curves-supp.tex
\begin{figure*}[t!]
\centering
\begin{subfigure}[c]{0.3\linewidth}
    \centering
    \includegraphics[width=1\textwidth]{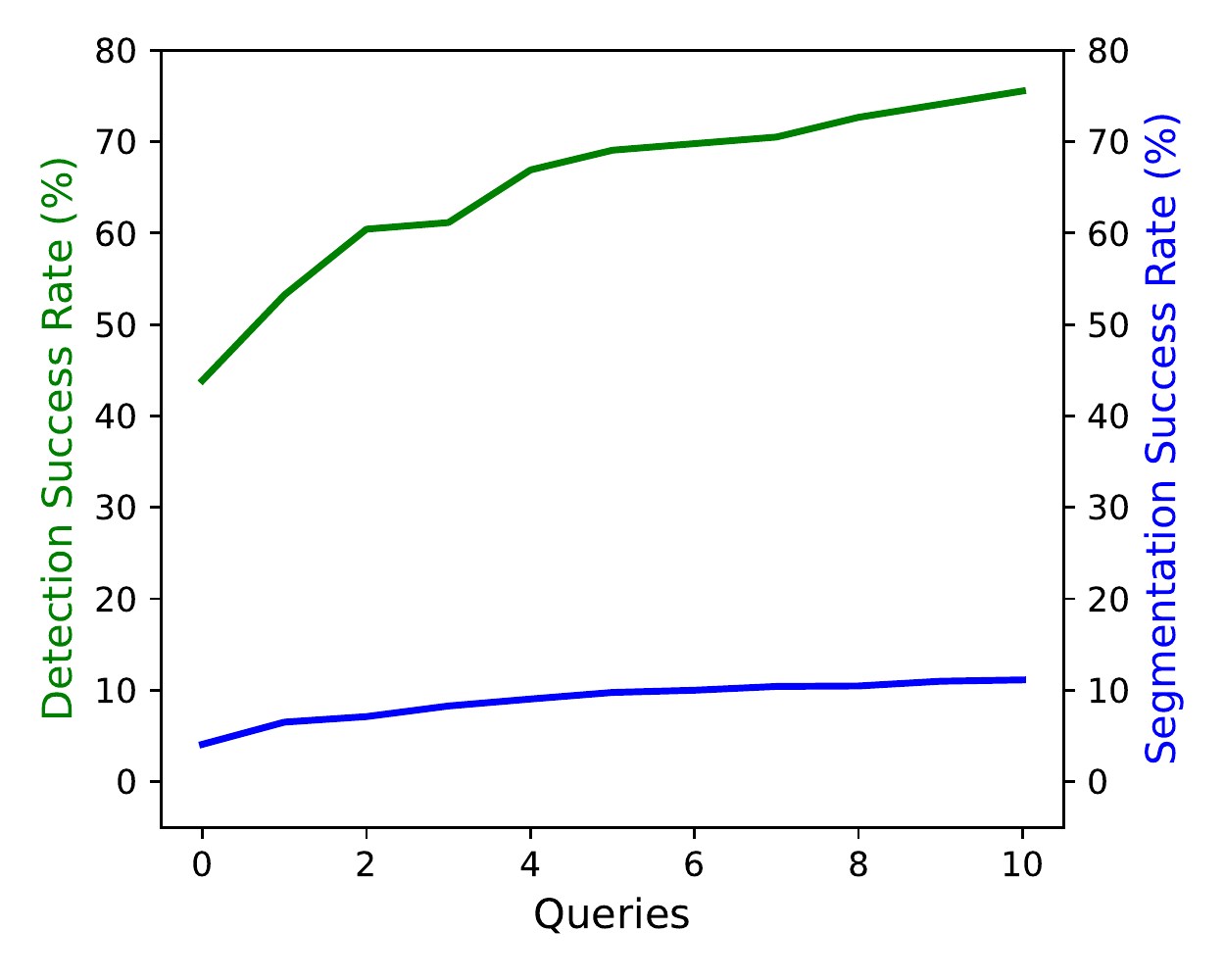}
    \caption{Attacks generated for object detection only, $\ell_\infty \leq 10$}
    \label{fig:sub-joint-det}
\end{subfigure}
~~
\begin{subfigure}[c]{0.3\linewidth}
    \centering
    \includegraphics[width=1\textwidth]{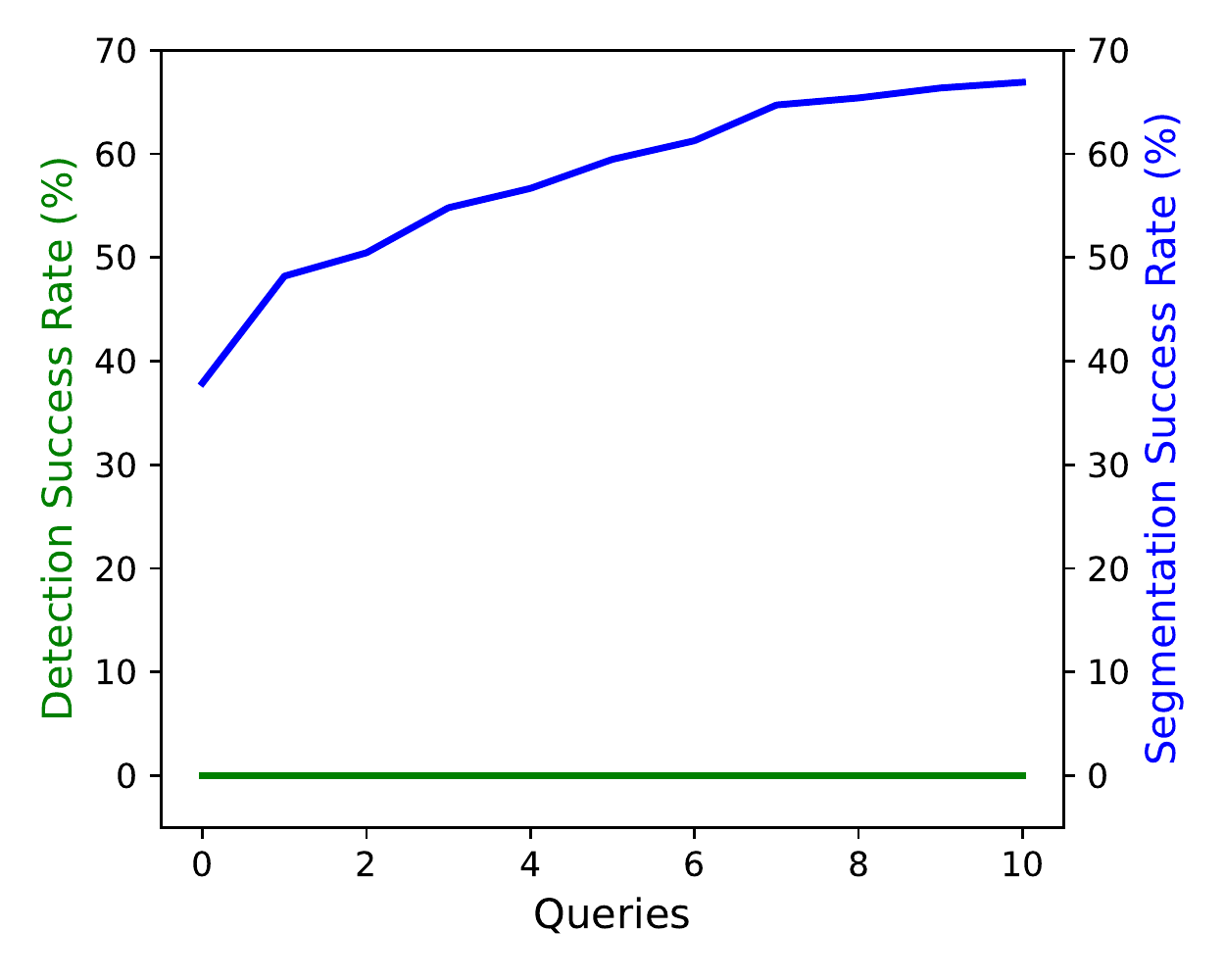}
    \caption{Attacks generated for semantic segmentation only, $\ell_\infty \leq 10$}
    \label{fig:sub-joint-seg}
\end{subfigure}
~~
\begin{subfigure}[c]{0.3\linewidth}
    \centering
    \includegraphics[width=1\textwidth]{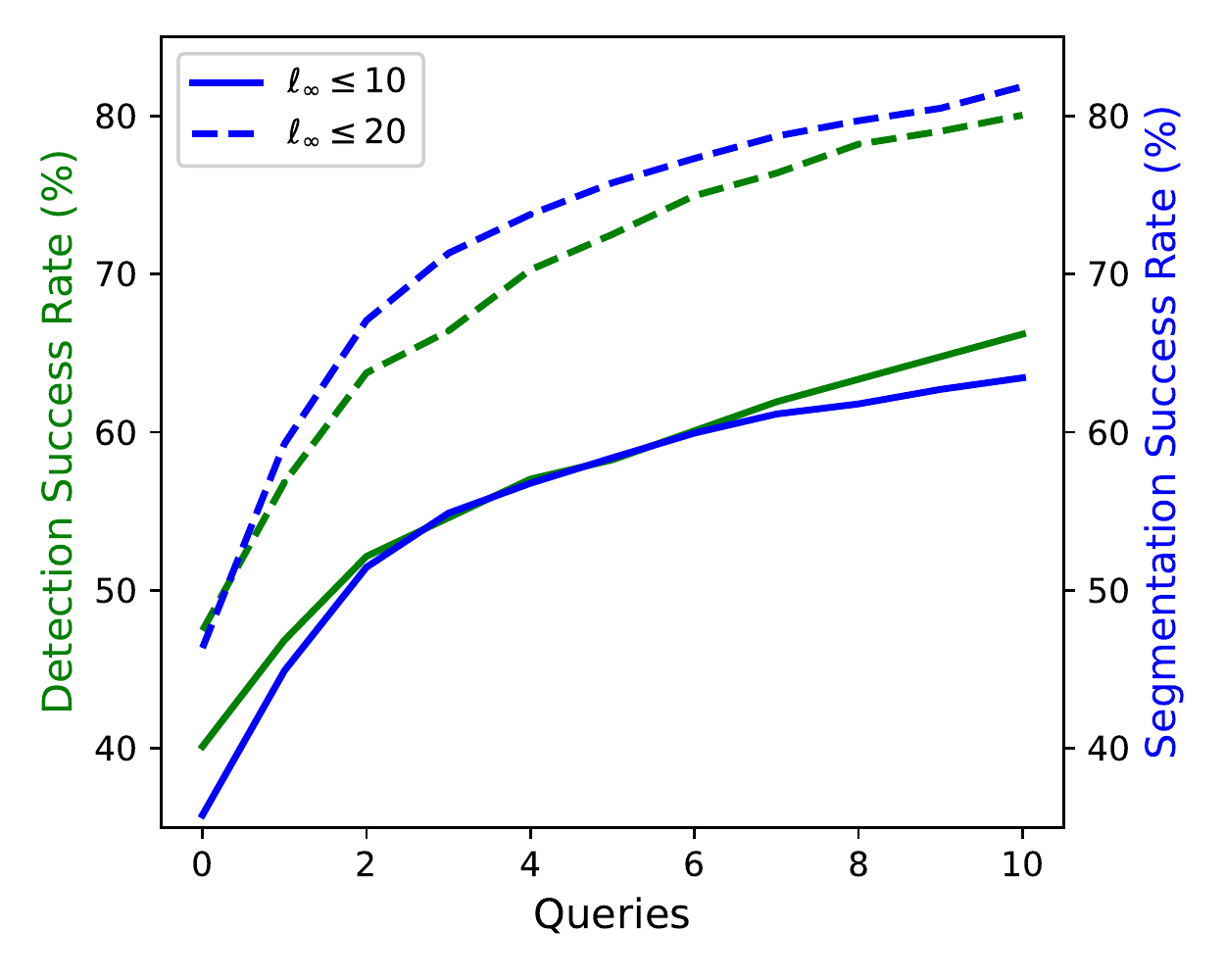}
    \caption{Attacks generated jointly for object detection and semantic segmentation, $\ell_\infty \leq \{10, 20\}$}
    \label{fig:sub-joint-all}
\end{subfigure}
\caption{Comparison between task-specific attacks and joint attack performance on blackbox object detector (\texttt{RetinaNet}) and segmentation model (\texttt{PSPNet}). Green curves denote attack success rate for object detectors, and blue curves denote pixel success rate for semantic segmentation. (a) Attacks generated with an object detector surrogate do not transfer for semantic segmentation. (b) Attacks generated with semantic segmentation models surrogate do not transfer for object detectors. (c) Attacks generated by a surrogate of object detectors and semantic segmentation (along with weight balancing and optimization) provide successful attacks for blackbox object detectors and semantic segmentation models. }
\label{fig:obj-joint-curves}
\end{figure*}

%% file: figures/fig-joint-attack.tex
\begin{figure*}[t!]
\centering
\begin{subfigure}[c]{0.9\linewidth}
    \centering
    \includegraphics[width=1\textwidth]{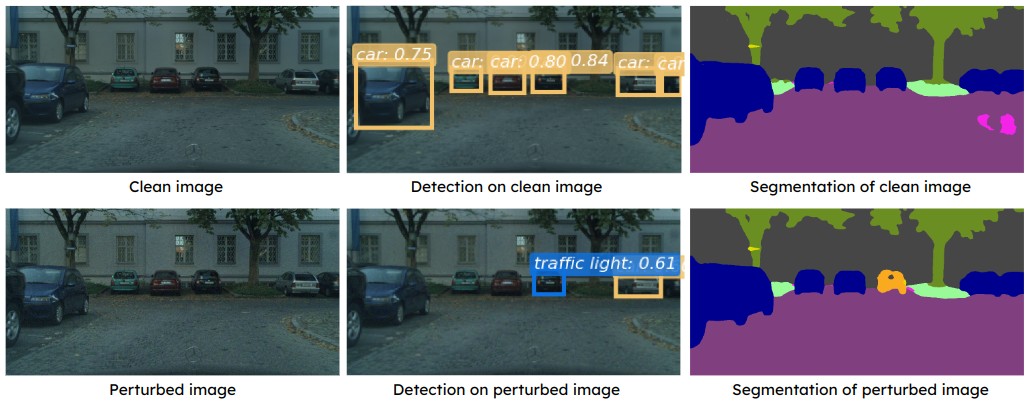}
    \caption{Our method can generate attacks to fool multiple blackbox object detector and blackbox semantic segmentation models jointly. First row shows a clean image from Cityscapes dataset, detection with \texttt{RetinaNet} and segmentation with \texttt{PSPNet}. Second row shows the perturbed image using ensemble surrogates \{\texttt{Faster RCNN}, \texttt{YOLOv3}, \texttt{FCN}, \texttt{UPerNet}\}, and detection and segmentation results on the perturbed image. We generate perturbation to map the Car in the middle to Traffic Light. Image id: \texttt{lindau\_000026\_000019}}
    \label{fig:obj-joint-attack}
\end{subfigure}
\\
\begin{subfigure}[c]{0.9\linewidth}
    \centering
    \includegraphics[width=1\textwidth]{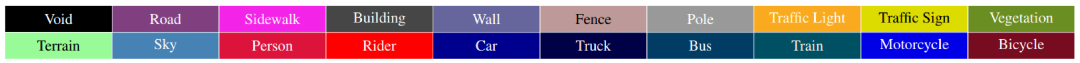}
    \caption{Color encoding for segmentation maps in CityScapes dataset}
    \label{fig:cityscapes-legends}
\end{subfigure}
\vspace{-5pt}
\caption{Visual adversarial examples of our method that generates successful attacks to fool a blackbox object detector and a blackbox semantic segmentation model using a single perturbed image. }
\label{fig:obj-joint-examples}
\vspace{-5mm} 
\end{figure*}

%% file: supp.tex
\newpage
\clearpage

\begin{strip} 
    \begin{center} 
        {\Large  \textbf{Ensemble-based Blackbox Attacks on Dense Prediction \\ (Supplementary Material)}}
    \end{center} 
\end{strip}

\newcommand{\beginsupplement}{%
        \setcounter{table}{0}
        \renewcommand{\thetable}{S\arabic{table}}%
        \setcounter{figure}{0}
        \renewcommand{\thefigure}{S\arabic{figure}}%
     }
\makeatletter
\newcommand\refwithdefault[2]{%
  \@ifundefined{r@#1}{%
    #2%
  }{%
    \ref{#1}%
  }%
}
\makeatother

\section*{Summary}
In the supplementary material, we provide additional results and analyses on joint attacks for multiple blackbox models of different dense prediction tasks, attacks on object detection, attacks on semantic segmentation, and corresponding visualization of adversarial examples for qualitative evaluation. We also report runtime and resource usage.

\appendix
\beginsupplement

\section{Joint attack for multiple blackbox models}
In this section, we provide additional visualization results
for joint (targeted) blackbox attacks against object detection and semantic segmentation models.

\input{figures/fig-joint-examples.tex}

\noindent \textbf{More Visualization of adversarial examples.} We visualize some adversarial examples in \cref{fig:obj-joint-examples}. In \cref{fig:joint-example1}, we show an example where our method generates a single perturbed image to map the bicycle on the right-hand-side to train. In object detection results we see the label for the Bicycle bounding box has been changed to Train, and for the segmentation map, the corresponding region has changed to teal color encoding for Train as well. In \cref{fig:joint-example2}, the generated perturbed image maps the car in the middle to traffic light. Note that the bounding box for the Car in the middle changed to Traffic Light for the object detector and the same area in the semantic segmentation map changed the color to orange (corresponding to Traffic Light label). 

\section{Attacks against object detection}

\input{tables/tab-obj-ablation-coco.tex}

\noindent \textbf{Attacks using different surrogate models.} In our previous experiments (\cref{tab:obj-ablation} and \cref{tab:obj-ablation-coco}), we follow the model selection in \cite{cai2022context} for a fair comparison. We can easily replace the surrogate models with different ones and expect its effectiveness across different settings.
For example, we can replace YOLOv3 with Deformable DETR (denoted as Deform) and get similar results, as shown in \cref{tab:obj-replaceYOLO} below. The experiment setup and victim  models are same as reported in \cref{tab:obj-replaceYOLO} for $\ell_\infty = 20$. 

\input{tables/tab-supp-obj-different-surrogates}

\noindent \textbf{Comparisons with zero-query attacks.} 
Zero-Query attack (ZQA) \cite{cai2022zero} does not rely on any feedback from the victim. It assesses the attack success probability on the surrogate model before launching a single and most promising attack against the victim. Due to these differences in problem setting, we do not directly compare with this method in the main paper. Here we compare the numbers reported from corresponding manuscripts in \cref{tab:obj-replaceYOLO}. ZQA uses a single surrogate model without any feedback from the victim model. It performs worse than the few-query attacks \cite{cai2022context} with 3--5 queries, and our method clearly outperforms both of them. 

\noindent \textbf{Comparison with conventional query-based attacks}. Existing query-based methods, including GARSDC \cite{liang2022large} and PRFA \cite{Liang21Parallel}, require thousands of queries (which is prohibitive) and they are only applicable for untargeted attacks.
Furthermore, their perturbations are clearly visible, see Fig. 5 in \cite{liang2022large}, while our perturbations remain imperceptible. 
\input{tables/tab-supp-obj-compare-conventional}
For these key differences, we did not include their comparison in the main paper, but here we provide a mAP score comparison with them. We use 5 surrogate models from~\cref{tab:obj-ensemble-size} and perform vanishing attacks on ATSS \cite{zhang2020bridging} model, we show in \cref{tab:compare-conventional} that our method can achieve a near-zero mAP within just a few queries ($Q$).

\section{Attacks against semantic segmentation}
\noindent \textbf{Attacks on Pascal VOC dataset.} We generate adversarial attacks using different sizes of ensemble and report mIoU scores on the Pascal VOC dataset in \cref{tab:seg-untar-voc}. Similar to the results on the Cityscapes dataset in~Tab.~\cref{tab:seg-untar-cs}, as we increase the number of surrogate models from 2 to 6, the attack performance improves (indicated by smaller mIoU scores). Attack performance of our method further improves with weight optimization (with $Q=20$). These results show that by adjusting the weights of the surrogate ensemble, we can improve the attack performance. Our attack method with $N=6$ surrogate models provides 27--29\% improvement in mIoU scores compared to DS attack for the victim models \texttt{PSPNet-Res50} and \texttt{DeepLabV3-Res50}. Note that DS attack uses these two models as the whitebox surrogates as well victim models.  In contrast, we keep all four victim models \texttt{PSPNet-Res50, DeepLabV3-Res50, PSPNet-Res101, DeepLabV3-Res101} out of our ensemble. 
Our surrogate ensemble consists of \texttt{FCN, UPerNet, PSANet, GCNe, ANN, EncNet} with \texttt{ResNet50} backbones, which reflects a more realistic setting where the victim blackbox model is different from any of the surrogate models.

\noindent \textbf{Effect of backbones on attack performance.} 
We note that for VOC dataset results in \cref{tab:seg-untar-voc}, our method provides high attack success for blackbox victim models with \texttt{ResNet50} backbone. However, the attack performance on victim models with \texttt{ResNet101} backbone degrades (as reflected by large mIoU values). To further demonstrate the effectiveness of our attack, we replace the backbones of the surrogate models with the \texttt{ResNet101} backbones while keeping the rest of model architectures same as the original ensemble. Results reported in \cref{tab:seg-untar-voc-r101} show that if we replace surrogate models with \texttt{ResNet101} backbones (same backbone as the victim blackbox models), then our attack method provides significantly better results. 

\noindent \textbf{Attack performance on different backbones.} 
We performed additional experiments using \texttt{FCN} and \texttt{PSPNet} methods and \texttt{MobileNetV2} and \texttt{ResNeSt} (denoted as -mv2 and -s101 in \cref{tab:seg-backbones}) backbones for victim models.
The attack setting corresponds to~Tab.~\cref{tab:seg-targeted}. Due to the great difference in backbones across surrogate and victim, the attack performance drops. Nevertheless, the attack performance improves significantly as we increase the ensemble size and optimize ensemble weights. Results are reported in \cref{tab:seg-backbones}.

\input{tables/tab-rebuttal-seg.tex}

\noindent \textbf{Attack performance on surrogate models.} For the sake of completeness, we also report attack performance on the whitebox surrogate models for both untargeted and targeted attacks in \cref{tab:seg-ensemble-size-untar} and \cref{tab:seg-ensemble-size-tar}. We observe that as we increase the number of models in the ensemble from $N = 1$ to $N = 5$, we can achieve better attack performance on all the whitebox and blackbox victim models we tested. Attacks that are successful on blackbox victim models are almost always successful on all surrogate models.

\input{tables/tab-seg-untar-voc.tex}
\input{tables/tab-seg-untar-voc-r101.tex}

\input{tables/tab-seg-untar-wb.tex}
\input{tables/tab-seg-tar-wb.tex}

\noindent \textbf{Visualization of adversarial examples.} We present some visual examples of untargeted attacks in \cref{fig:seg-untar-examples} and targeted attacks in \cref{fig:seg-tar-examples}. We observe that the attacks generated by surrogate model do not transfer to the victim model for untargeted or targeted cases (i.e., $Q=0$). The attacks generated after weight optimization (i.e., $Q=20$) succeed for untargeted and targeted attacks. 
Our targeted attack setup is visually explained in \cref{fig:seg-tgt-explain}. Instead of mapping every pixel prediction to an arbitrary target label, we focus on attacking a single object $y$ in the original prediction (e.g.``road'' in \cref{fig:tgt-original} with white bounding-box). We select the target label  $y^\star$ as the class that appears most frequently as the least-likely label of the pixels in the selected region. For example, \cref{fig:tgt-ll} shows class ``building'' in grey color as the least likely class in the target region. Finally, we generate attack to replace the entire selected region in the original prediction to its target label (\cref{fig:tgt-target}).

\input{figures/fig-seg-vis-untar.tex}
\input{figures/fig-seg-tgt-explain.tex}

\input{figures/fig-seg-vis-tar.tex}
 
\section{Runtime and resource usage}
We performed experiments on a single RTX 3090 GPU. Average time per query to attack an object detector for a $375\times500$  image with ensemble size $N=\{2,5\}$ is $\{0.5,1\}$sec. Average time per query to attack a segmentation model for a $512\times1024$ image with ensemble size $N=\{2,5\}$ is $\{2.5,5.5\}$sec.

%% file: figures/fig-joint-examples.tex
\begin{figure*}[t!]
\centering
\begin{subfigure}[c]{0.9\linewidth}
    \centering
    \includegraphics[width=1\textwidth]{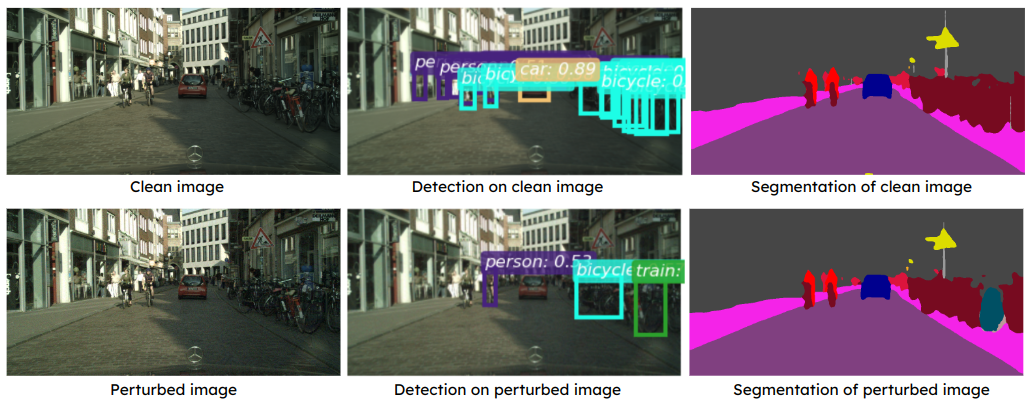}
    \caption{Generate perturbation to map the Bicycle on the right-hand-side to Train. Image id: \texttt{munster\_000140\_000019}}
    \label{fig:joint-example1}
\end{subfigure}
~~
\begin{subfigure}[c]{0.9\linewidth}
    \centering
    \includegraphics[width=1\textwidth]{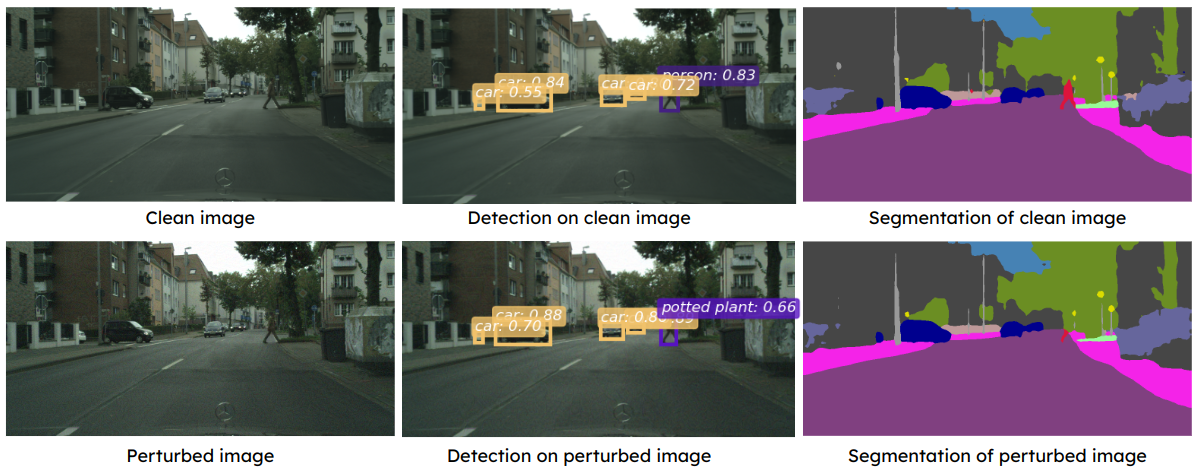}
    \caption{Generate perturbation to map the Pedestrian to Potted Plant. Image id: \texttt{munster\_000006\_000019}}
    \label{fig:joint-example2}
\end{subfigure}
~~
\begin{subfigure}[c]{0.9\linewidth}
    \centering
    \includegraphics[width=1\textwidth]{figures/joint-figs/fig-joint-legends.png}
    \caption{Color encoding for segmentation maps in CityScapes dataset}
    \label{fig:cityscapes-legends}
\end{subfigure}
\caption{Visual adversarial examples of our method that generates successful attacks to fool a blackbox object detector and a blackbox semantic segmentation model using a single perturbed image. }
\label{fig:obj-joint-examples}
\end{figure*}

%% file: tables/tab-obj-ablation-coco.tex
\begin{table*}[]

\centering
\small
\caption{Targeted attack success rate (\%) for different methods on COCO dataset. Similar setting as in~\cref{tab:obj-ablation}.}
\label{tab:obj-ablation-coco}

\begin{tabular}{ccccc|ccccc}
\hline
\multirow{2}{*}{\begin{tabular}[c]{@{}c@{}}Perturbation \\ Budget\end{tabular}} & \multirow{2}{*}{\begin{tabular}[c]{@{}c@{}}Weight \\ Balancing\end{tabular}} & \multirow{2}{*}{\begin{tabular}[c]{@{}c@{}}Weight \\ Optimization\end{tabular}} & \multicolumn{2}{c|}{\textbf{Surrogate Ensemble}} & \multicolumn{5}{c}{\textbf{Blackbox Victim Models} (ASR $\uparrow$)} \\
 &  &  & FRCNN & YOLOv3 & RetinaNet & Libra & Fovea & Free & DETR \\ \hline
\multirow{5}{*}{$\ell_{\infty} = 10$} & \xmark & \xmark & 19.6 & 79.7 & 4.6 & 5.0 & 4.4 & 6.6 & 2.6 \\
 & \xmark & \cmark & 49.8 & \textbf{97.8} & 13.5 & 16.2 & 14.4 & 22.6 & 8.4 \\
 & \cmark & \xmark & 57.2 & 65.3 & 16.2 & 17.6 & 16.6 & 24.0 & 5.4 \\
 & \cmark & \cmark & \textbf{78.0} & 86.1 & \textbf{31.7} & \textbf{32.0} & \textbf{32.3} & \textbf{41.6} & \textbf{15.4} \\
 & \multicolumn{2}{c}{Context-aware Attack \cite{cai2022context}} & 41.2 & 54.4 & 12.0 & 11.2 & 18.6 & 25.0 & 10.8 \\ \hline
\multirow{5}{*}{$\ell_{\infty} = 20$} & \xmark & \xmark & 25.8 & 82.2 & 8.9 & 9.8 & 8.4 & 13.2 & 5.6 \\
 & \xmark & \cmark & 62.4 & \textbf{98.2} & 23.0 & 32.2 & 22.4 & 32.2 & 13.2 \\
 & \cmark & \xmark & 68.8 & 75.8 & 25.5 & 28.0 & 27.1 & 38.0 & 13.8 \\
 & \cmark & \cmark & \textbf{88.9} & 94.5 & \textbf{48.5} & \textbf{53.8} & \textbf{49.5} & \textbf{65.6} & \textbf{31.0} \\
 & \multicolumn{2}{c}{Context-aware Attack \cite{cai2022context}} & 64.4 & 70.0 & 20.8 & 22.2 & 35.4 & 40.8 & 20.0 \\ \hline
\multirow{5}{*}{$\ell_{\infty} = 30$} & \xmark & \xmark & 29.0 & 82.2 & 8.8 & 9.4 & 13.3 & 14.6 & 6.4 \\
 & \xmark & \cmark & 69.0 & \textbf{99.3} & 27.9 & 34.6 & 31.2 & 43.6 & 17.6 \\
 & \cmark & \xmark & 72.7 & 78.6 & 32.5 & 33.8 & 34.1 & 41.6 & 14.8 \\
 & \cmark & \cmark & \textbf{91.7} & 95.5 & \textbf{57.6} & \textbf{64.4} & \textbf{58.3} & \textbf{71.2} & \textbf{36.6} \\
 & \multicolumn{2}{c}{Context-aware Attack \cite{cai2022context}} & 68.6 & 75.4 & 27.2 & 27.2 & 39.2 & 46.2 & 21.2 \\ \hline
\end{tabular}

\end{table*}

%% file: tables/tab-supp-obj-different-surrogates.tex
\begin{table*}[h]
\centering
\caption{Replacing YOLOv3 with Deformable DETR. Correspond to~\cref{tab:obj-ablation}, perturbation budget $\ell_{\infty} = 20$.} \label{tab:obj-replaceYOLO}
\small
\begin{tabular}{cccc|ccccc}
\hline
\multirow{2}{*}{\begin{tabular}[c]{@{}c@{}}Weight \\ Balancing\end{tabular}} & \multirow{2}{*}{\begin{tabular}[c]{@{}c@{}}Weight \\ Optimization\end{tabular}} & \multicolumn{2}{c|}{\textbf{Surrogate Ensemble}} & \multicolumn{5}{c}{\textbf{Blackbox Victim Models} (ASR $\uparrow$)} \\
 &  & FRCNN & Deform & Retina & Libra & Fovea & Free & DETR \\ \hline
\xmark & \xmark & 8.5 & 69.5 & 10.6 & 4.0 & 8.0 & 10.5 & 12.0 \\
\xmark & \cmark & 34.6 & 94.3 & 36.7 & 25.0 & 33.5 & 53.0 & 38.5 \\
\cmark & \xmark & 68.0 & 80.5 & 47.2 & 38.5 & 37.5 & 57.5 & 26.5 \\
\cmark & \cmark & \textbf{88.1} & \textbf{95.0} & \textbf{74.9} & \textbf{70.5} & \textbf{73.0} & \textbf{84.0} & \textbf{56.0} \\
\multicolumn{2}{c}{ZQA \cite{cai2022zero}} & 88.2 & - & 44.0 & 51.4 & 53.4 & - & - \\
\hline
\end{tabular}
\end{table*}

%% file: tables/tab-supp-obj-compare-conventional.tex
\begin{table}[h!]
\centering
\small
\caption{Comparison with conventional query-based attacks.}
\label{tab:compare-conventional}
\begin{tabular}{lcc}
\hline
\multirow{2}{*}{Method} & \multicolumn{2}{c}{ATSS\cite{zhang2020bridging}} \\
 & mAP $\downarrow$ & $Q$ $\downarrow$ \\ \hline
Clean & 0.54 & N/A \\
PRFA \cite{Liang21Parallel} & 0.20 & 3500 \\
GARSDC \cite{liang2022large} & 0.04 & 1837 \\
Ours & \textbf{0.00} & \textbf{10} \\ \hline
\end{tabular}
\end{table}

%% file: tables/tab-rebuttal-seg.tex
\begin{table}[h]
\small
\centering
\caption{Semantic segmentation targeted pixel success ratio (PSR) (\%) for blackbox victim models with different backbones.}
\label{tab:seg-backbones}
\begin{tabular}{c|c|cccc}
\hline
\multirow{2}{*}{$Q$}    & \multirow{2}{*}{$N$} & \multicolumn{4}{c}{\textbf{Blackbox Victim Models} (PSR $\uparrow$)}                        \\
                      &                    & FCN-mv2        & FCN-s101       & PSP-mv2     & PSP-s101    \\ \hline
\multirow{3}{*}{0}    & 1                  & 33.26          & 1.01           & 3.96           & 2.71           \\
                      & 3                  & 30.39          & 1.39           & 5.82           & 6.94           \\
                      & 5                  & 38.92          & 3.12           & 7.84           & 8.32           \\ \hline
\multirow{2}{*}{20}   
                      & 3                  & 50.31          & 22.79          & 24.09          & 54.06          \\
                      & 5                  & \textbf{53.09}          & \textbf{34.57} & \textbf{30.20}          & \textbf{60.43}          \\ \hline
\end{tabular}

\end{table}

%% file: tables/tab-seg-untar-voc.tex
\begin{table*}[h]
\centering
\small
\caption{mIoU scores (\%) for untargeted attacks on semantic segmentation models with  Pascal VOC dataset. The lower value indicates better attack performance. Surrogate of ensemble sizes $N=2,4,6$. We compare $Q=0$ (i.e. direct transfer attack) with $Q=20$ ensemble attack performance. Results show enabling the ensemble query introduced attack performance increments. \textcolor{blue}{Blue} numbers represent whitebox attacks.}
\label{tab:seg-untar-voc}

\begin{tabular}{c|c|cccc}
\hline
 &  & \multicolumn{4}{c}{\textbf{Blackbox Victim Models} (mIoU $\downarrow$)} \\
\multirow{-2}{*}{Method} & \multirow{-2}{*}{\textbf{Whitebox Surrogate}} & PSPNet-Res50 & PSPNet-Res101 & DeepLabV3-Res50 & DeepLabV3-Res101 \\ \hline
Clean Images & - & 76.78 & 78.47 & 76.17 & 78.70 \\ \hline
 & PSPNet-Res50 & \textcolor{blue}{5.09} & 37.06 & 6.57 & 38.98 \\
\multirow{-2}{*}{Baseline} & DeepLabV3-Res50 & 3.63 & 22.01 & \textcolor{blue}{3.14} & 22.58 \\ \hline
\cellcolor[HTML]{FFFFFF} & PSPNet-Res50 & \textcolor{blue}{2.07} & 16.10 & 2.56 & 18.57 \\
\multirow{-2}{*}{\cellcolor[HTML]{FFFFFF}DS} & DeepLabV3-Res50 & 2.31 & \textbf{12.32} & \textcolor{blue}{2.15} & \textbf{13.64} \\ \hline
 & $N=2$ & 14.33 & 35.47 & 12.31 & 35.31 \\
 & $N=4$ & 8.74 & 29.41 & 7.92 & 28.01 \\
\multirow{-3}{*}{Ours ($Q = 0$)} & $N=6$ & 7.28 & 24.28 & 6.75 & 24.63 \\ \hline
 & $N=2$ & 5.56 & 27.49 & 4.43 & 28.46 \\
 & $N=4$ & 2.23 & 22.24 & 2.09 & 20.34 \\
\multirow{-3}{*}{Ours ($Q = 20$)} & $N=6$ & \textbf{1.69} & 18.07 & \textbf{1.53} & 17.61 \\ \hline
\end{tabular}

\end{table*}

%% file: tables/tab-seg-untar-voc-r101.tex
\begin{table*}[h]
\centering
\small
\caption{mIoU scores (\%) for untargeted attacks on semantic segmentation models with  Pascal VOC dataset. The lower value indicates better attack performance. Surrogate of ensemble sizes $N=1$ to $6$.  
We compare the performance of \texttt{ResNet50} and \texttt{ResNet101} backbones in the ensemble. The attack performance on \texttt{ResNet101} backbone victim models increases if we use the surrogate models with \texttt{ResNet101} backbone. Note there is no weight optimization for $N=1$.}
\label{tab:seg-untar-voc-r101}

\begin{tabular}{c|c|cccc}
\hline
 &  & \multicolumn{2}{c}{} & \multicolumn{2}{c}{} \\
 &  & \multicolumn{2}{c}{\multirow{-2}{*}{\textbf{Blackbox Victim: PSPNet-Res101} (mIoU $\downarrow$)}} & \multicolumn{2}{c}{\multirow{-2}{*}{\textbf{Blackbox Victim: DeeplabV3-Res101} (mIoU $\downarrow$)}} \\ \cline{3-6} 
\multirow{-3}{*}{$Q$} & \multirow{-3}{*}{$N$} & Ensemble backbone: Res50 & \multicolumn{1}{c|}{Ensemble backbone: Res101} & Ensemble backbone: Res50 & Ensemble backbone: Res101 \\ \hline
 & 1 & 38.51 & \multicolumn{1}{c|}{24.76} & 38.66 & 25.98 \\
 & 2 & 35.47 & \multicolumn{1}{c|}{21.50} & 35.31 & 21.54 \\
 & 3 & 31.95 & \multicolumn{1}{c|}{17.65} & 32.39 & 18.02 \\
 & 4 & 29.41 & \multicolumn{1}{c|}{14.53} & 28.01 & 14.32 \\
 & 5 & 25.82 & \multicolumn{1}{c|}{13.67} & 24.79 & 12.28 \\
\multirow{-6}{*}{0} & \cellcolor[HTML]{FFFFFF}6 & 24.28 & \multicolumn{1}{c|}{12.49} & 24.63 & 12.35 \\ \hline
\cellcolor[HTML]{FFFFFF} & 2 & 27.49 & \multicolumn{1}{c|}{8.78} & 28.46 & 8.80 \\
\cellcolor[HTML]{FFFFFF} & 3 & 24.80 & \multicolumn{1}{c|}{5.15} & 22.55 & 5.69 \\
\cellcolor[HTML]{FFFFFF} & 4 & 22.24 & \multicolumn{1}{c|}{5.49} & 20.34 & 4.49 \\
\cellcolor[HTML]{FFFFFF} & 5 & 19.62 & \multicolumn{1}{c|}{\textbf{3.27}} & 18.31 & \textbf{3.13} \\
\multirow{-5}{*}{\cellcolor[HTML]{FFFFFF}20} & 6 & \textbf{18.07} & \multicolumn{1}{c|}{4.04} & \textbf{17.61} & 3.32 \\ \hline
\end{tabular}

\end{table*}

%% file: tables/tab-seg-untar-wb.tex
\begin{table*}[h]

\centering
\small
\caption{Semantic segmentation untargeted attack mIoU scores (\%) for blackbox victim models and whitebox surrogate models with different ensemble sizes ($N$). The lower value indicates better attack performance. Experiment with CityScapes dataset, $\ell_\infty \leq 8$. PSP-r50, PSP-r101, DL3-r50, DL3-r101 stands for \texttt{PSPNet} and \texttt{DeepLabV3} built on \texttt{ResNet50}, \texttt{ResNet101} backbone respectively.}
\label{tab:seg-ensemble-size-untar}

\begin{tabular}{ccccccc|cccc}
\hline
\multirow{2}{*}{$N$} & \multicolumn{6}{c|}{\textbf{Surrogate Ensemble}} & \multicolumn{4}{c}{\textbf{Blackbox Victim Models} (mIoU $\downarrow$)} \\
 & FCN & UPerNet & PSANet & GCNet & ANN & EncNet & PSP-r50 & PSP-r101 & DL3-r50 & DL3-r101 \\ \hline
1 & 2.42 & - & - & - & - & - & 2.68 & 6.92 & 5.16 & 10.13 \\
2 & 1.28 & 1.06 & - & - & - & - & 1.38 & 2.88 & 1.15 & 3.50 \\
3 & 1.45 & 1.06 & 1.05 & - & - & - & 1.13 & 2.39 & 0.95 & 2.67 \\
4 & 1.25 & 0.97 & 0.87 & 0.91 & - & - & \textbf{0.79} & 2.04 & \textbf{0.73} & 1.80 \\
5 & 1.18 & 0.96 & 0.93 & 0.91 & 1.14 & - & 0.78 & 1.69 & 0.89 & 2.09 \\
6 & 1.26 & 1.08 & 1.08 & 1.05 & 1.25 & 1.16 & 0.90 & \textbf{1.55} & 0.94 & \textbf{1.09} \\ \hline
\end{tabular}

\end{table*}

%% file: tables/tab-seg-tar-wb.tex
\begin{table*}[h]

\centering
\small
\caption{Semantic segmentation targeted pixel success ratio (PSR) (\%) for blackbox victim models and whitebox surrogate models with different ensemble sizes ($N$). The higher value indicates better attack performance. Experiment with CityScapes dataset, $\ell_\infty \leq 8$. PSP-r50, PSP-r101, DL3-r50, DL3-r101 stands for \texttt{PSPNet} and \texttt{DeepLabV3} built on \texttt{ResNet50}, \texttt{ResNet101} backbone respectively.}
\label{tab:seg-ensemble-size-tar}

\begin{tabular}{cccccc|cccc}
\hline
\multirow{2}{*}{$N$} & \multicolumn{5}{c|}{\textbf{Surrogate Ensemble}} & \multicolumn{4}{c}{\textbf{Blackbox Victim Models} (PSR $\uparrow$)} \\
 & FCN & UPerNet & PSANet & GCNet & ANN & PSP-r50 & PSP-r101 & DL3-r50 & DL3-r101 \\ \hline
1 & 69.51 & - & - & - & - & 39.15 & 10.21 & 35.02 & 7.58 \\
2 & 84.62 & 89.30 & - & - & - & 83.97 & 51.80 & 82.70 & 46.95 \\
3 & 79.64 & 85.48 & 82.89 & - & - & 88.88 & 64.63 & 85.55 & 60.88 \\
4 & 83.82 & 88.50 & 87.00 & 88.12 & - & 91.51 & 64.28 & 87.19 & 63.88 \\
5 & 86.55 & 91.10 & 89.75 & 90.00 & 87.82 & \textbf{92.91} & \textbf{69.09} & \textbf{88.95} & \textbf{69.65} \\ \hline
\end{tabular}

\end{table*}

%% file: figures/fig-seg-vis-untar.tex
\begin{figure*}[t!]
\centering
\begin{subfigure}[c]{0.9\linewidth}
    \centering
    \includegraphics[width=1\textwidth]{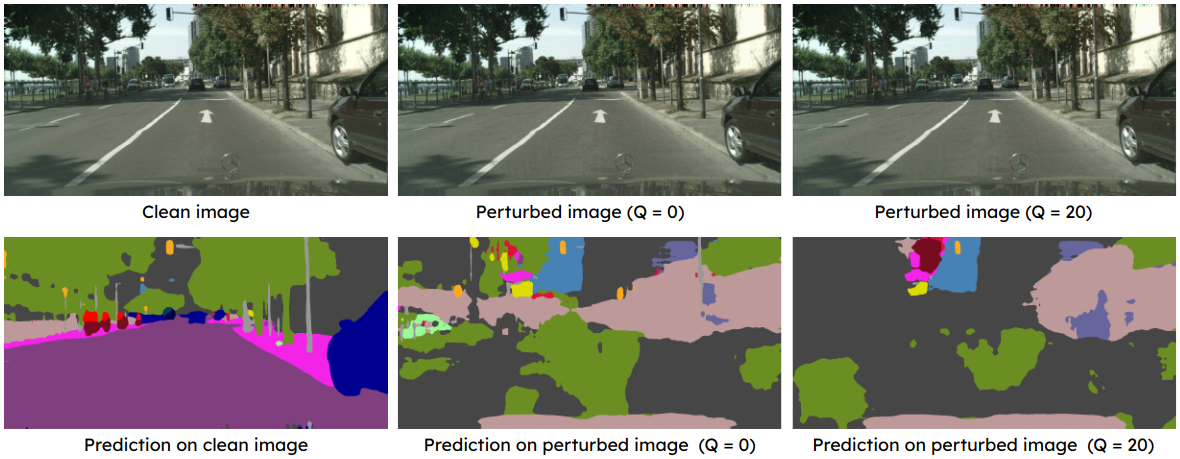}
    \caption{Generate perturbation to maximize prediction error.
    We use the $N=2$ surrogate ensemble to generate this attack against blackbox victim \texttt{PSPNet-Res50}.
    Image id: \texttt{frankfurt\_000001\_005703}}
    \label{fig:seg-untar-example1}
\end{subfigure}
~~
\begin{subfigure}[c]{0.9\linewidth}
    \centering
    \includegraphics[width=1\textwidth]{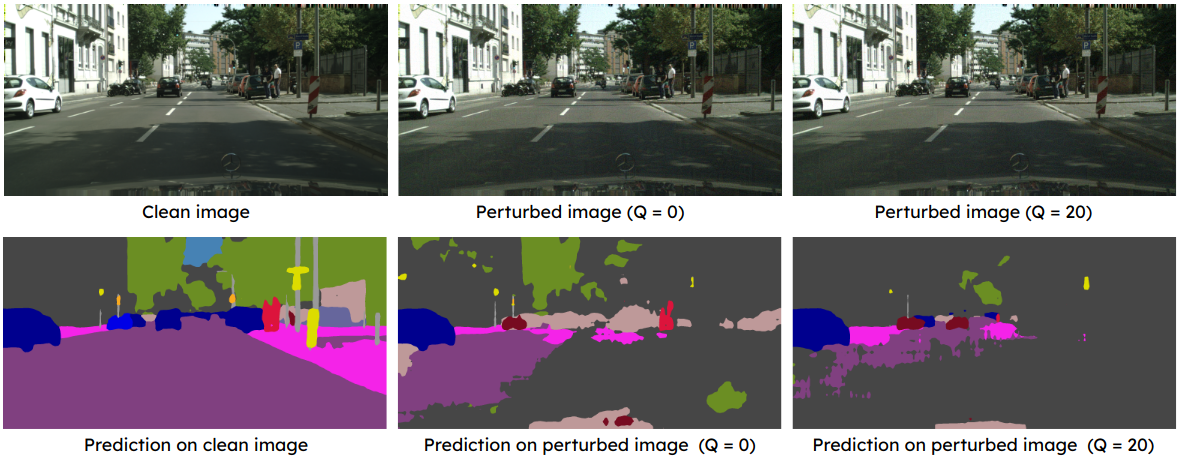}
    \caption{Generate perturbation to maximize prediction error.
    We use the $N=2$ surrogate ensemble to generate this attack against blackbox victim \texttt{DeepLabV3-Res101}. Image id: \texttt{frankfurt\_000000\_022797}}
    \label{seg-untar-example2}
\end{subfigure}

\caption{Visual adversarial examples of our method for untargeted attacks to fool a blackbox semantic segmentation model. }
\label{fig:seg-untar-examples}
\end{figure*}

%% file: figures/fig-seg-tgt-explain.tex
\begin{figure*}[h]
\centering
\begin{subfigure}[c]{0.3\linewidth}
    \centering
    \includegraphics[width=1\textwidth]{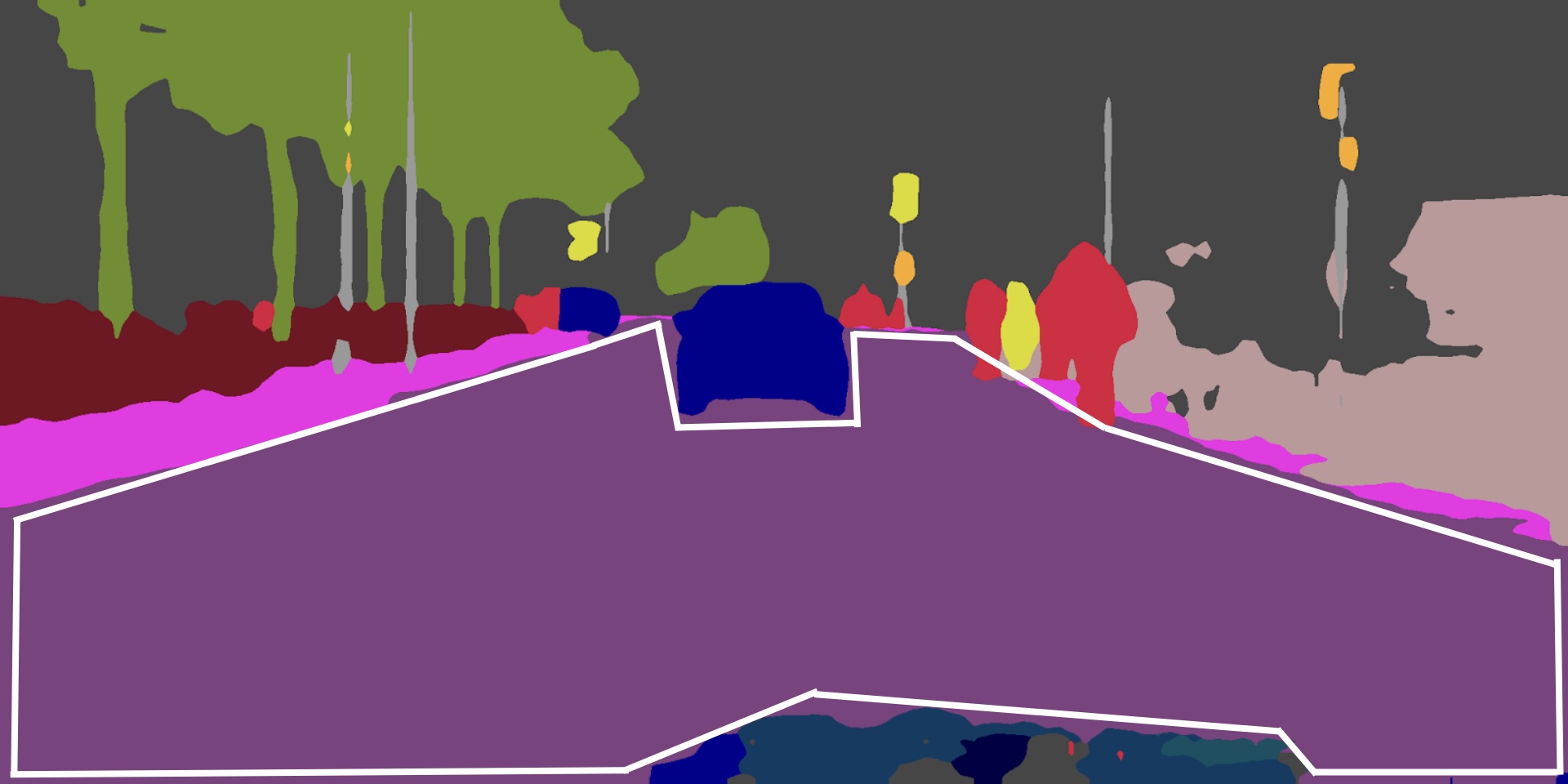}
    \caption{Original Prediction}
    \label{fig:tgt-original}
\end{subfigure}
~~
\begin{subfigure}[c]{0.3\linewidth}
    \centering
    \includegraphics[width=1\textwidth]{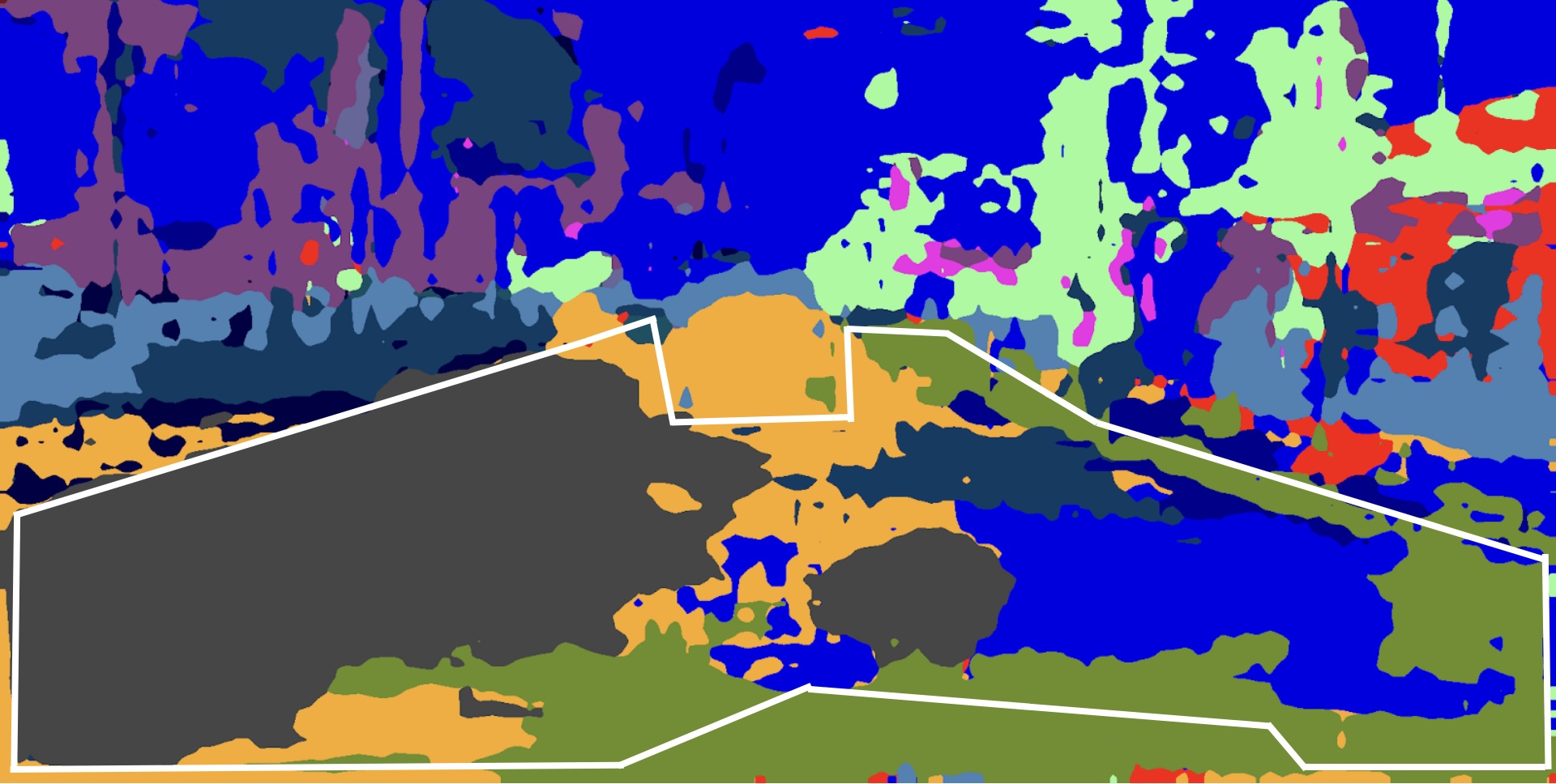}
    \caption{Least-likely Prediction}
    \label{fig:tgt-ll}
\end{subfigure}
~~
\begin{subfigure}[c]{0.3\linewidth}
    \centering
    \includegraphics[width=1\textwidth]{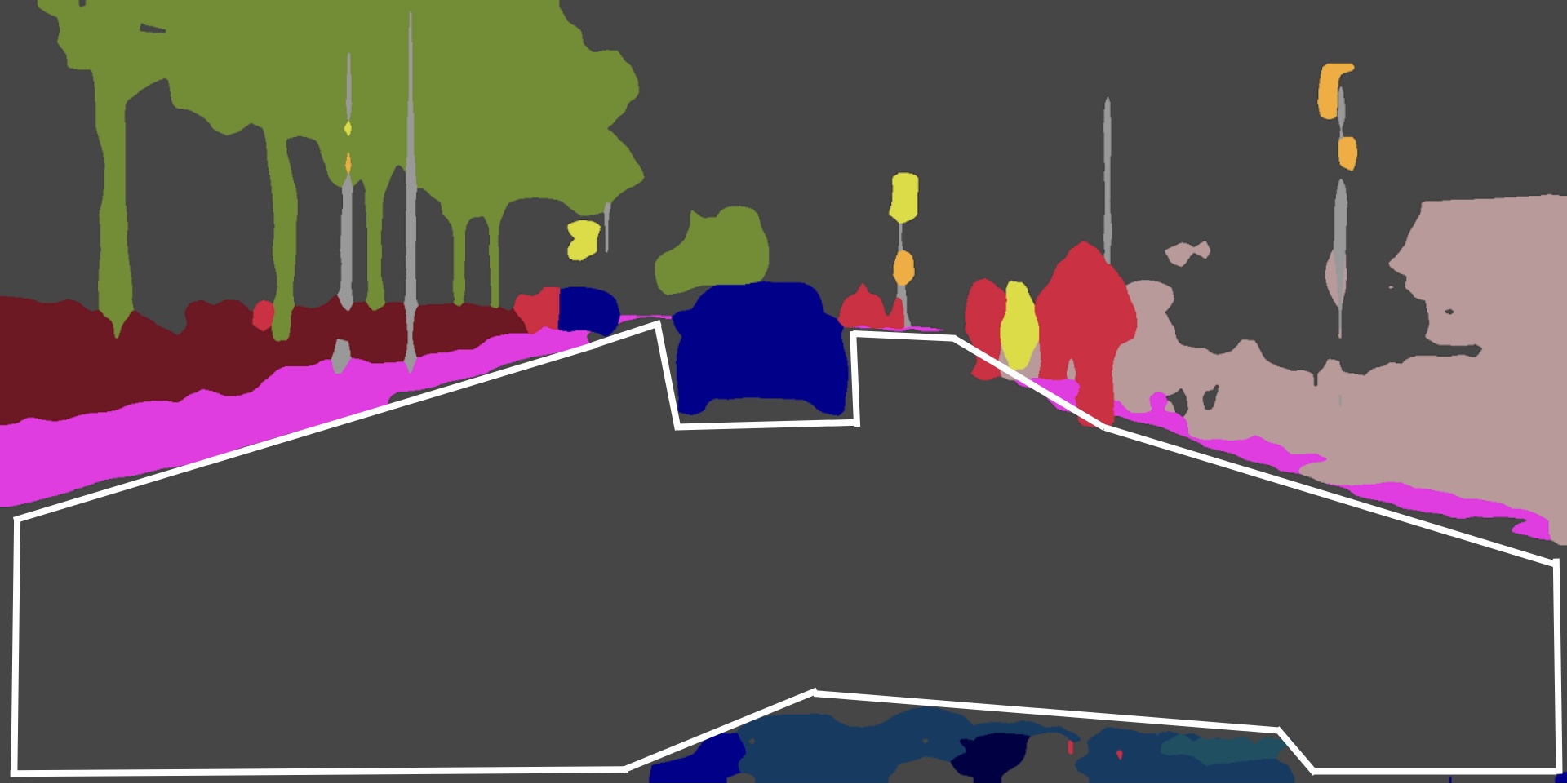}
    \caption{Attack Target}
    \label{fig:tgt-target}
\end{subfigure}
\caption{Our segmentation targeted attack setup. We select an object region $y$ in the original prediction from surrogate \texttt{FCN} (\cref{fig:tgt-original}). Identify the targeted label $y^\star$ from \cref{fig:tgt-ll} and craft the attack target \cref{fig:tgt-target}. 
Image id: \texttt{frankfurt\_000001\_007857}}

\label{fig:seg-tgt-explain}
\end{figure*}

%% file: figures/fig-seg-vis-tar.tex
\begin{figure*}[t!]
\centering
\begin{subfigure}[c]{0.9\linewidth}
    \centering
    \includegraphics[width=1\textwidth]{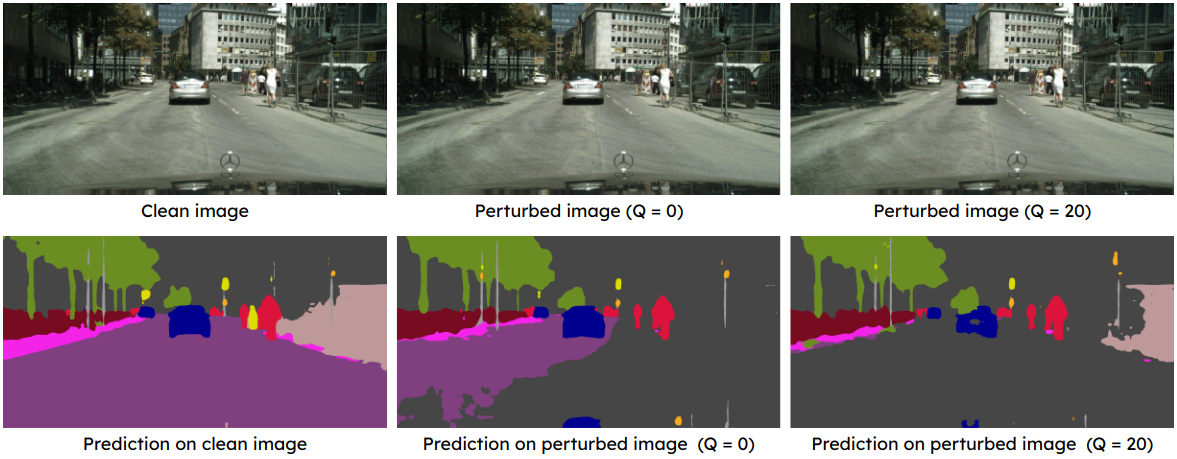}
    \caption{Generate perturbation to map the Road region to Building.
    We use the $N=2$ surrogate ensemble to generate this attack against blackbox victim \texttt{PSPNet-Res50}.
    For the direct transfer attack (at $Q=0$), only $67.08\%$ of the pixels of the target region are successfully mapped to the desired class. After weight optimization (with $Q=20$), pixel success rate increases to $99.77\%$.
    Image id: \texttt{frankfurt\_000001\_007857}}
    \label{fig:seg-tar-example1}
\end{subfigure}
~~
\begin{subfigure}[c]{0.9\linewidth}
    \centering
    \includegraphics[width=1\textwidth]{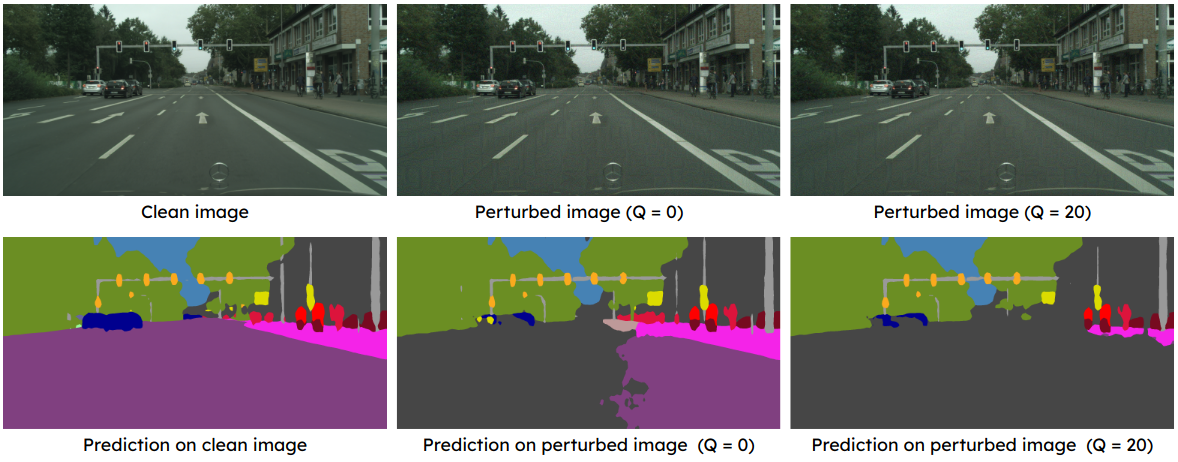}
    \caption{Generate perturbation to map the Road region to Building.
    We use the $N=6$ surrogate ensemble to generate this attack against blackbox victim \texttt{DeepLabV3-Res50}.
     For the direct transfer attack (at $Q=0$), only $63.06\%$ of the pixels of the target region are successfully mapped to the desired class. After weight optimization (with $Q=20$), pixel success rate increases to $99.98\%$.
    Image id: \texttt{munster\_000003\_000019}}
    \label{fig:seg-tar-example2}
\end{subfigure}

\caption{Visual adversarial examples of our method for targeted attacks to fool a blackbox semantic segmentation model. }
\label{fig:seg-tar-examples}
\end{figure*}